%% file: camera_ready.tex
\documentclass[10pt,twocolumn,letterpaper]{article}

\usepackage{iccv}              % To produce the CAMERA-READY version
\newcommand{\methodname}{\mbox{Gaussian Atlas}}
\newcommand{\datasetname}{\mbox{GaussianVerse}}

\usepackage[table]{xcolor}
\definecolor{iccvblue}{rgb}{0.21,0.49,0.74}
\usepackage[pagebackref,breaklinks,colorlinks,allcolors=iccvblue]{hyperref}
\usepackage{graphicx}
\usepackage{mathtools}
\usepackage{pifont}

\usepackage{pdflscape}
\usepackage{makecell}

\def\paperID{234} % *** Enter the Paper ID here
\def\confName{ICCV}
\def\confYear{2025}

\definecolor{best_color}{HTML}{FCE5CD}
\definecolor{better_color}{HTML}{DEEDF2}

\title{Repurposing 2D Diffusion Models with Gaussian Atlas for 3D Generation}

\author{Tiange Xiang$^{1,2}$\thanks{Part of this work was done while Tiange Xiang was an intern at Meta Reality Labs, under the mentorship of Kai Li and Chengjiang Long.}, Kai Li$^{2}$\thanks{Corresponding authors.}, Chengjiang Long$^{2}$, Christian Häne$^{2}$, Peihong Guo$^{2}$, \\
Scott Delp$^{1}$, Ehsan Adeli$^{1}$, Li Fei-Fei$^{1}$\footnotemark[2]\\
$^{1}$Stanford University, $^{2}$Meta Reality Labs\\
\textit{{\Large\href{https://cs.stanford.edu/~xtiange/projects/gaussianatlas}{\textcolor{purple}{https://cs.stanford.edu/~xtiange/projects/gaussianatlas}}}\vspace{-1em}}
}

\begin{document}
\maketitle
\input{sec/0_abstract}    
\input{sec/1_intro}

\input{sec/2_relatedwork}

\input{sec/3_dataset}

\input{sec/4_methods}

\input{sec/5_experiment}

\input{sec/6_conclusion}
{
    \small
    \bibliographystyle{ieeenat_fullname}
    \bibliography{main}
}

\input{sec/X_suppl}

\end{document}

%% file: sec/0_abstract.tex
\begin{abstract} 
% Text-to-image diffusion models have seen significant development recently due to increasing availability of paired 2D data. While a similar trend is emerging in 3D generation, the scarcity of high-quality 3D data has resulted in less competitive 3D diffusion models compared to their 2D counterparts. In this work, we show how 2D diffusion models, originally trained for text-to-image generation, can be repurposed for 3D object generation. We introduce \methodname, a representation of 3D Gaussians with dense 2D grids, which enables the fine-tuning of 2D diffusion models for generating 3D Gaussians. Our approach shows a successful transfer learning from a pre-trained 2D diffusion model to 2D manifold flattened from 3D structures. To facilitate model training, a large-scale dataset, \datasetname, is compiled to comprise 205K high-quality 3D Gaussian fittings of a diverse array of 3D objects. Our experiment results indicate that text-to-image diffusion models can also serve as 3D content generators.

Recent advances in text-to-image diffusion models have been driven by the increasing availability of paired 2D data. However, the development of 3D diffusion models has been hindered by the scarcity of high-quality 3D data, resulting in less competitive performance compared to their 2D counterparts. To address this challenge, we propose repurposing pre-trained 2D diffusion models for 3D object generation. We introduce \methodname, a novel representation that utilizes dense 2D grids, enabling the fine-tuning of 2D diffusion models to generate 3D Gaussians. Our approach demonstrates successful transfer learning from a pre-trained 2D diffusion model to a 2D manifold flattened from 3D structures. To support model training, we compile \datasetname, a large-scale dataset comprising 205K high-quality 3D Gaussian fittings of various 3D objects. Our experimental results show that text-to-image diffusion models can be effectively adapted for 3D content generation, bridging the gap between 2D and 3D modeling.
\end{abstract}

%% file: sec/1_intro.tex
\section{Introduction}
\label{sec:intro}
Understanding the 3D world is crucial for numerous real-world applications. In this work, we focus on generative 3D modeling with the primary objective of generating high-quality 3D assets from given textural descriptions \cite{jain2022zero, poole2022dreamfusion}.

\begin{figure}
    \centering
    \includegraphics[width=0.99\linewidth]{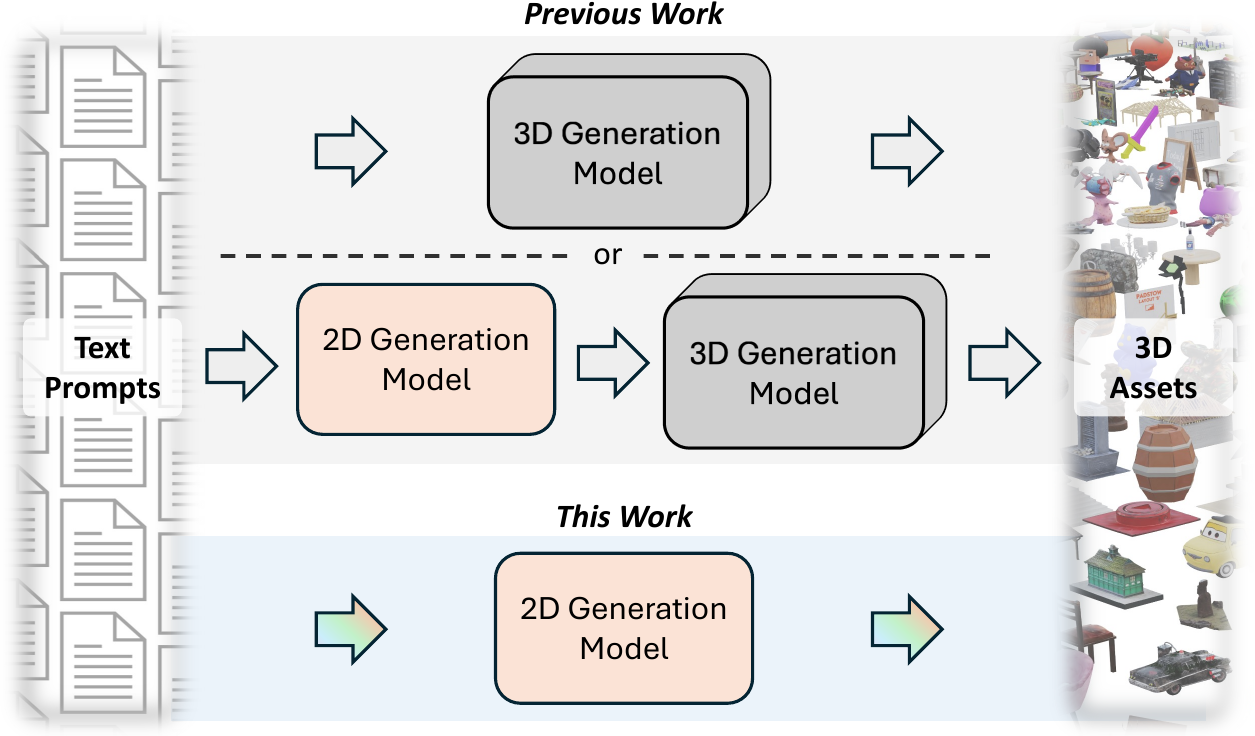}
    \vspace{-0.5em}
    % \caption{Previous work built either 2D diffusion models for text to image generation \cite{ho2020denoising, Rombach_2022_CVPR} or 3D diffusion models for text to object generation \cite{zhang2024gaussiancube, he2024gvgen}. In this work, we repurpose pretrained 2D diffusion models for text to 3D generations.}
    \caption{Previous 3D generators are either end-to-end 3D models \cite{zhang2024gaussiancube, he2024gvgen} or a combination of a multi-view 2D generator followed by a 2D-to-3D lifting model \cite{xu2024grm, wang2024crm}. In this work, we achieve 3D object generation by directly fine-tuning 2D generation models.}
    \vspace{-0.5em}
    \label{fig:intro}
\end{figure}

% In this paper, we propose to use 2D diffusion models to generate 3D objects via 2D parameterizations of 3D Gaussians.

With advances in 3D content representation, diffusion models~\cite{ho2020denoising,song2020score} begin to play a predominant role in 3D generation \cite{luo2021diffusion, nichol2022point, hong2023lrm}. Among different 3D representations, 3D Gaussians \cite{3dgs} have gained popularity for modeling 3D scenes and objects due to their high efficiency, explicit structure, and superior rendering quality. Notably, training diffusion models to generate 3D Gaussians requires high-quality ``ground truth" 3D Gaussians, which must be pre-fitted to the 3D objects~\cite{he2024gvgen,tang2023dreamgaussian,ma2024shapesplat}. 

To advance the community, we constructed a large-scale dataset named \textbf{GaussianVerse}, comprising high-quality 3D Gaussian fittings for a diverse range of objects. Unlike previous studies that also pre-compute 3D Gaussian references, GaussianVerse provides higher-quality fittings with the minimum required number of Gaussians, through a novel pruning strategy and a more effective, but compute-intensive, fitting process.

Designing standalone diffusion models for 3D generation is straightforward. However, such models have significant limitations especially when trained solely on 3D data, as high-quality 3D data is relatively scarce compared to 2D images. To achieve higher rendering fidelity in 3D generation, it is desirable to leverage the learned prior knowledge from well-pretrained 2D diffusion models \cite{poole2022dreamfusion, elizarov2024geometry, liu2024pi3d, mercier2024hexagen3d}. Some methods have been proposed to incorporate pretrained 2D diffusion models with frozen weights into the 3D generation pipeline through complex designs, such as score-distilled sampling \cite{poole2022dreamfusion} and collaborative controls \cite{elizarov2024geometry}. 

%Although these existing works show some promising results, they are still far from being perfect to generate high-quality 3D objects even with very simple text inputs.

In this paper, we propose a fresh perspective that repurposes 2D diffusion models for 3D generation through direct fine-tuning. To fully harness the capabilities of these 2D diffusion models, we introduce {\bf Gaussian Atlas}, a novel 2D representation of 3D Gaussians. This representation enables direct fine-tuning by first transporting unorganized 3D Gaussians into a standard 3D sphere and then applying equirectangular projection to map them into a square 2D grid, creating a Gaussian Atlas. We show that these Gaussian atlases facilitate transfer of the prior knowledge captured in 2D diffusion models to the 3D generation task. By doing so, our approach provides a means to leverage the learned 2D priors for 3D generation, unlocking new possibilities for efficient and effective 3D content creation. 
%For reliable fine-tuning of 2D diffusion models, we further prepared a wide range of 3D Gaussians representation of 3D objects through compute-intensive per-object offline fitting.

%We shall emphasize that GIMDiffusion \cite{elizarov2024geometry} explores a similar approach for 3D object generation with pretrained 2D diffusion models. Their method leverages an existing image-based representation of 3D objects \cite{gu2002geometry} and trains 2D diffusion models to generate geometry images and albedo maps in the UV space, ultimately producing a textured mesh. However, their generated UV maps are sparse, leading to artifacts at the seams on the assembled 3D mesh. In contrast, our method generates 3D Gaussians by introducing a dense 2D representation. % As another contribution, we computed and will release a large-scale 3D Gaussian object dataset, providing a valuable resource for the broader community.

To summarize, our major contributions are three-fold: (\emph{i}) We present a large-scale dataset, \textbf{\datasetname}, consisting of 205,737 high-quality 3D Gaussian fittings for diverse objects sampled from Sketchfab \cite{sketchfab}; (\emph{ii}) We propose a novel 2D representation of 3D Gaussians, \textbf{\methodname}, facilitating 3D generation tasks through a new perspective by fine-tuning pretrained 2D diffusion models; (\emph{iii}) We demonstrate that the proposed approaches surpass state-of-the-art 3D Gaussian generators in terms of both generation-prompt alignment and user preferences.

%% file: sec/2_relatedwork.tex
\section{Related Work}
\label{sec:relatedwork}

\noindent\textbf{Repurposing 2D diffusion models.} Diffusion models \cite{ho2020denoising, Rombach_2022_CVPR, esser2024scaling}, are originally designed for text-to-image generation. When trained on large-scale internet images, prior knowledge learned from diverse domains can be transferred to different tasks. Marigold \cite{ke2024repurposing} is an example of how fine-tuning pretrained Latent Diffusion (LD) models \cite{Rombach_2022_CVPR} can benefit depth prediction. Marigold first re-parameterizes single-channel depth maps as RGB images and uses the pretrained VAE encoder to compress both the input image as well as its paired depth map into a shared latent space. The two latents are then concatenated to enable conditional generation of depth latents, which are subsequently decoded back into RGB space through the VAE decoder. More recent work, such as GeoWizard \cite{fu2025geowizard} and GenPercept \cite{xu2024diffusion}, explored the repurposing of LD models for more geometric prediction tasks and beyond. As more and more studies have shown the potential of finetuning LD for different purposes, this work is among the first to explore how a LD model can be adapted for 3D generation, addressing the significant disparity between 3D objects and 2D images.

 \begin{figure*}[ht]
    \vspace{-0.5em}
    \centering
    \includegraphics[width=0.9\linewidth]{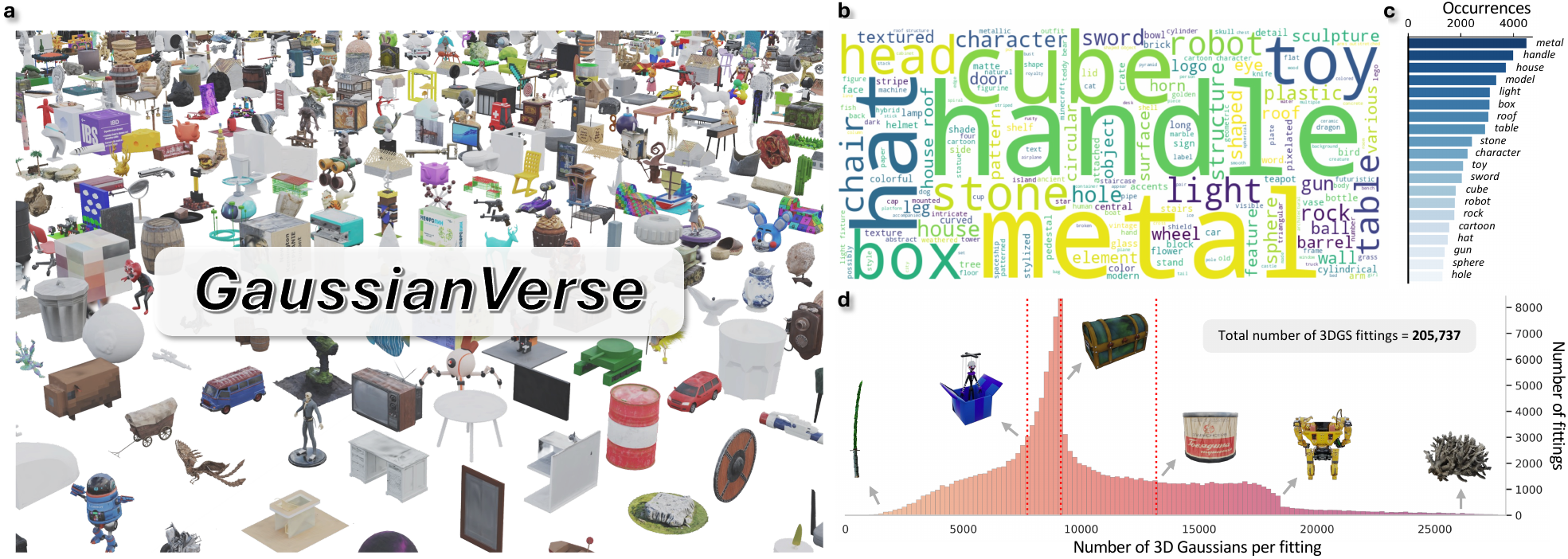}
    \vspace{-0.5em}
    % \caption{\textbf{\datasetname\ offers high-quality 3DGS fittings for a diverse range of 3D objects.} (a) We compare between our proposed 3DGS fitting method, the densification-constrained fitting method introduced by GaussianCube \cite{zhang2024gaussiancube}, and the upper bound method, Scaffold-GS \cite{lu2024scaffold}. \emph{Our method achieves high fitting quality comparable to the upper bound with significantly fewer 3D Gaussians of positive opacity.} (b) A word cloud and (c) a list of the most frequently occurring words generated from the captions \cite{luo2023scalable} of the fitted objects. (d) The distribution of the number of valid Gaussians per fitting shows that our method generates fewer Gaussians for simple objects and more for complex ones. Red dotted lines indicate the 25\%, 50\%, and 75\% percentile of the distribution.}
    \caption{(a) \textbf{\datasetname}\ offers high-quality 3DGS fittings for diverse 3D objects. (b) A word cloud and (c) a list of the most frequently occurring words generated from the captions \cite{luo2023scalable} of the fitted objects. (d) The distribution of the number of Gaussians per fitting shows that our method generates fewer Gaussians for simple objects and more for complex ones. Red dotted lines indicate the 25\%, 50\%, and 75\% percentile of the distribution.}
    \vspace{-0.5em}
    \label{fig:GO}
\end{figure*}

% \noindent\textbf{Diffusion models for 3D Gaussian generation.}
\vspace{0.5em}\noindent\textbf{3D generation with Gaussian splatting.} Unlike 2D images, 3D objects are more difficult to generate due to the additional dimension and geometric constraints \cite{mildenhall2021nerf, 3dgs}. Among all 3D generation methods \cite{tochilkin2024triposr, ren2024xcube, lan2025ln3diff, chen20243dtopia, wang2024crm, xu2024instantmesh, li2023instant3d}, there is a line of work that utilizes diffusion models to generate 3D Gaussians, which is closely related to this paper. GSD \cite{mu2024gsd} introduced rendering guidance to constrain the sampling of 3D Gaussians with 2D observations. L3DG \cite{roessle2024l3dg} is another example, which embeds 3D Gaussians onto a dense latent grid and learns a diffusion model in the latent space for generation. Moreover, GVGen \cite{he2024gvgen} and GaussianCube \cite{zhang2024gaussiancube} achieve 3D Gaussian diffusion by transforming sparsely located Gaussians into more structured 3D volumes. GVGen fits 3D Gaussians directly on a volume with offsets, while GaussianCube applies Optimal Transport to move 3D Gaussians to vertices of a predefined 3D grid. Gaussian Anything~\cite{lan2024gaussiananything} adopts a two-step approach that generates Gaussians by auto-encoding point cloud latents. Recently, TRELLIS~\cite{xiang2025structured} designed a structured representation that supports decoding to various 3D formats. In this work, we propose to transform 3D Gaussians onto 2D atlases, to leverage the power of well pretrained 2D diffusion models and unify the architecture of 2D and 3D generation.

\vspace{0.5em}\noindent\textbf{2D representations of 3D content.} Representing 3D objects as 2D planes is not new. A major line of research encodes 3D structures using implicit Triplanes \cite{chan2022efficient}. NFD \cite{shue20233d} was one of the first attempts to train 2D diffusion models to generate Triplanes, from which neural fields can be derived to reconstruct 3D objects. To enhance 3D coherence, CRM \cite{wang2024crm} integrates multi-view observations into the Triplane generation process. InstantMesh \cite{xu2024instantmesh} and Instant3D \cite{li2023instant3d} follow a similar design but employ more sophisticated models to improve Triplane synthesis. Instead of generating neural fields, TriplaneGaussian \cite{zou2024triplane} decodes 3D Gaussian attributes directly from generated Triplanes. More recently, DiffGS \cite{zhou2024diffgs} has refined the TriplaneGaussian paradigm by learning a mapping between pre-fitted 3D Gaussians and Triplanes via a latent diffusion model. Several works share motivations similar to ours. For example, PI3D \cite{liu2024pi3d} fine-tunes a pretrained text-to-image generation model to produce Triplanes represented as “pseudo-images”, while HexaGen3D~\cite{mercier2024hexagen3d} leverages pretrained image diffusion models for improved Triplane generation. However, their fine-tuned diffusion models exhibit limited capacity, and an additional minutes-long refinement process is typically required to enhance the 3D generation.

In addition to Triplane-based representations, Splatter Image~\cite{szymanowicz2024splatter} trains a 2D network to infer pixel-wise Gaussian attributes from single images; Omages~\cite{yan2024object} trains 2D diffusion models to fit 2D UV maps of the geometry and materials of 3D objects, ultimately reconstructing a textured 3D mesh. However, Omegas supports only class-conditioned generations of a limited pre-defined categories;  GIMDiffusion~\cite{elizarov2024geometry} adopts a similar approach by representing the surface of 3D objects with geometry images \cite{gu2002geometry} and leveraging 2D diffusion models in the UV space; DiffSplat~\cite{lin2025diffsplat} trains VAE to compress multi-view 3D Gaussians into 2D latents and achieve multi-view generation through a pretrained 2D diffusion model. In contrast, our method introduces a more dense 2D representation of 3D Gaussians, yielding a novel approach for repurposing pretrained 2D diffusion models.

%% file: sec/3_dataset.tex
\section{\datasetname} \label{sec:dataset}
In this section, we present \datasetname, a large-scale dataset containing high-quality 3D Gaussian fittings for a wide range of 3D objects. A summary of this dataset is available in \tableautorefname~\ref{tab:3dgs_data} and \figureautorefname~\ref{fig:GO}. We describe the process of fitting per-object 3D Gaussians below. \emph{More details are provided in the supplementary materials.}

\begin{table}[t]
    \centering
    \resizebox{\linewidth}{!}{%
        \begin{tabular}{c|c|c}
            \toprule
            \# Gaussians per fitting & Total \# fittings & Compute time \\
            \hline
            10,435$\pm$4,453 & 205,737 & $>$3.8 GPU Years\\
            \bottomrule
        \end{tabular}
    }
    \vspace{-0.5em}
    \caption{\textbf{Statistics for the proposed \datasetname\ dataset.} We report the average (with standard deviation) number of 3D Gaussians per fitting, the total number of fitted objects, and the overall compute time spent on A100 GPUs.}
    \vspace{-0.5em}
    \label{tab:3dgs_data}
\end{table}

\noindent\textbf{Preliminaries: 3D Gaussian splatting \cite{3dgs}.} Given multi-view observations of a 3D object, 3D Gaussian Splatting (3DGS) represents the scene as a set of Gaussian distributions defined in 3D space. Each Gaussian is characterized by five attributes: its 3D location $\mathbf{x_i}\in\mathcal{R}^{3}$, color information $\mathbf{c_i}\in\mathcal{R}^{3}$, opacity $\mathbf{o}_i\in\mathcal{R}^{1}$, scale $\mathbf{s}_i\in\mathcal{R}^{3}$, and 3D rotation $\mathbf{r}_i\in\mathcal{R}^{3}$. The 3D covariance matrix $\Sigma_{i}$ can be represented as $\Sigma_{i}=\mathbf{r}_i\mathbf{s}_i\mathbf{r}_i^{T}$. A 2D image $C$ can be rendered from properly structured 3D Gaussians through $\alpha$-blending:
\begin{equation}
    C^\pi = \sum_{j=1} \mathbf{c}_j \omega_j^\pi \prod_{k=1}^{j-1} (1 - \omega_k^\pi),
\end{equation}
where $\pi$ is the camera pose, $\omega_i^\pi$ represents $z$-depth ordered opacity, computed as $\mathbf{o}_i e^{-\frac{1}{2} (\mathbf{x} - \mathbf{x}^\pi_i)^T (\Sigma_i^\pi)^{-1} (\mathbf{x} - \mathbf{x}^\pi_i)}$. The terms $(\mathbf{x}_i^\pi, \Sigma_i^\pi)$ denote the 2D Gaussian projection on the image plane derived from the 3D Gaussians with $\pi$.

\begin{figure*}[t]
    \centering
    \includegraphics[width=0.9\linewidth]{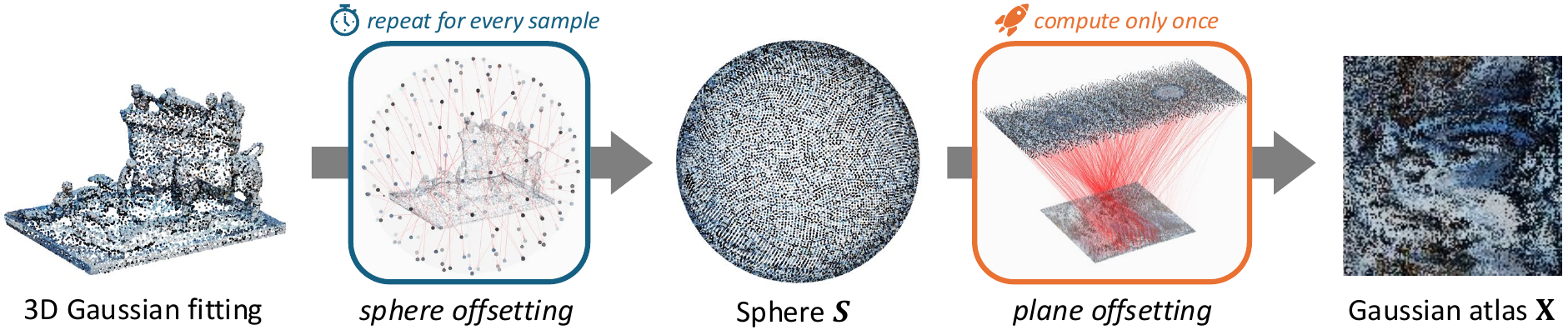}
    \vspace{-0.5em}
    \caption{\textbf{Representing 3D Gaussians as 2D Gaussian Atlas.} For each fitting, 3D Gaussians are first translated to the surface of a standard sphere $S$ via optimal transport. Then, the surface Gaussians are flattened onto the 2D plane via equirectangular projection with reusable indices. We obtain \methodname\ by reorganizing the flattened coordinates to pixels of a dense 2D square of size $\sqrt{N}\times\sqrt{N}$.}
    \vspace{-0.5em}
    \label{fig:atlas}
\end{figure*}

\vspace{0.5em}\noindent\textbf{3D Gaussian fitting with bounded quantity.} For training diffusion models, high-quality 3D Gaussian ``ground truth'' need to be pre-fitted for 3D objects in different formats. We build our fitting model upon the state-of-the-art, Scaffold-GS \cite{lu2024scaffold}, along with non-trivial modifications. First, we exclude view properties from the MLP predictors for attribute querying to enable more view-invariant applications. Second, we propose constraining the number of Gaussians per fitting, as also suggested in \cite{mu2024gsd, zhang2024gaussiancube}.

% \paragraph{3D Gaussian fitting with bounded quantity.} For training diffusion models, high quality 3D Gaussians 'ground truth' need be pre-fitted on a per-object basis. We build our fitting model upon the state-of-the-art, Scaffold-GS \cite{lu2024scaffold} with non-trivial modifications. First, we exclude view properties for MLP predictors for attributes querying, to enable more potential view-invariant applications. Then, we propose to constrain the number of Gaussians per fitting as also suggested in \cite{mu2024gsd, zhang2024gaussiancube}.

% We base our 3D Gaussian fittings on the state-of-the-art method, Scaffold-GS \cite{lu2024scaffold}. To enable view-invariant applications (e.g., diffusion model training), we exclude view properties from the inputs to MLP predictors used to query attributes for each 3D Gaussian.

Different 3D objects can exhibit diverse geometries and appearances, leading to highly variable 3DGS fittings. A common approach is to have a constant number of Gaussians per fitting \cite{mu2024gsd, zhang2024gaussiancube}, which enforces a uniform distribution of number of Gaussians across all 3D objects. Consequently, low-poly, simple objects can get excessively parameterized, with low, though not zero, scale and opacity values. Instead of constraining the number of Gaussians to be constant, we introduce the \textbf{visibility ranking} strategy that \emph{bounds} the number of fitted Gaussians at $\tau$, a self-defined parameter. Specifically, during 3DGS fitting, we track each Gaussian’s opacity to assess its visibility from randomly selected camera views. If the number of Gaussians exceeds the bound $\tau$ after densification, we sort their opacities and prune those with the lowest values. This pruning strategy differs from the original implementation \cite{3dgs}, which only prunes \emph{completely invisible} Gaussians. Additionally, to align with Scaffold-GS's design, we apply the same ranking and pruning based on the tracked visibility of anchors. Final Gaussians for pruning are selected by averaging the ranks from both tracking metrics.

Our visibility ranking strategy introduces minimal computational overhead while providing significant flexibility for various applications. See \figureautorefname~\ref{fig:dataset_comp} in the supplementary materials for our fitting results on objects with varying levels of detail and \figureautorefname~\ref{fig:GO} (d) for the distribution of the number of Gaussians across all fittings in \datasetname.

We optimize per-object 3D Gaussians by minimizing photometric losses against multi-view RGB renderings:
\begin{equation} \label{eq:3dgs_fit}
    \lambda_{rgb}'\mathcal{L}_{rgb} + \lambda_{ssim}'\mathcal{L}_{ssim} + \lambda_{lpips}'\mathcal{L}_{lpips} + \lambda_{reg}'\mathcal{R},
\end{equation}
where $\mathcal{L}_{rgb}$ represents the color space L1 loss, $\mathcal{L}_{ssim}$ is the negated structural similarity index, $\mathcal{L}_{lpips}$ is the perceptual loss, and $\mathcal{R}$ is a scaling regularization term \cite{lu2024scaffold}. The $\lambda$'s are weights for each loss. Note that the addition of perceptual loss nearly doubles the training time, but it is necessary for high rendering fidelity with fewer Gaussians.

\noindent\textbf{Dataset details.} \datasetname\ consists of \textbf{205,737} 3DGS fittings of 3D objects sampled from Sketchfab \cite{sketchfab} under CC licenses. We used the multi-view 2D renderings provided in \cite{zuo2024high} to perform per-object fittings. The 2D renderings came in 10 different categories, and we used renderings from 9 of the categories, excluding those of 'poor quality'. This curated list results in a wide coverage of diverse objects. We visualize a word cloud as well as the top 20 most frequent words from the captions \cite{luo2023scalable} in \figureautorefname~\ref{fig:GO} (b) and (c). For optimal efficiency and quality balance \cite{zhang2024gaussiancube,he2024gvgen}, we set the bound \(\tau\) to \(192 \times 192 = 36,864\). A cluster of A100 GPUs is employed for large-scale parallel fitting. Each 3DGS fitting job converges at around 20,000 steps, translating to approximately 10 minutes of fitting time per object, with a total of \textbf{over 3.8 A100 GPU years} spent.

\vspace{0.5em}\noindent\textbf{Comparison with related work.} Compared with several previous studies \cite{mu2024gsd, he2024gvgen, zhang2024gaussiancube} which also fit per-object 3D Gaussians for training diffusion models, we achieve higher-quality 3DGS fittings with fewer valid Gaussians but increased compute costs with a refined pruning strategy and a more effective, but compute-intensive, fitting process.

%% file: sec/4_methods.tex
\begin{figure*}[t]
    \centering
    \includegraphics[width=0.9\linewidth]{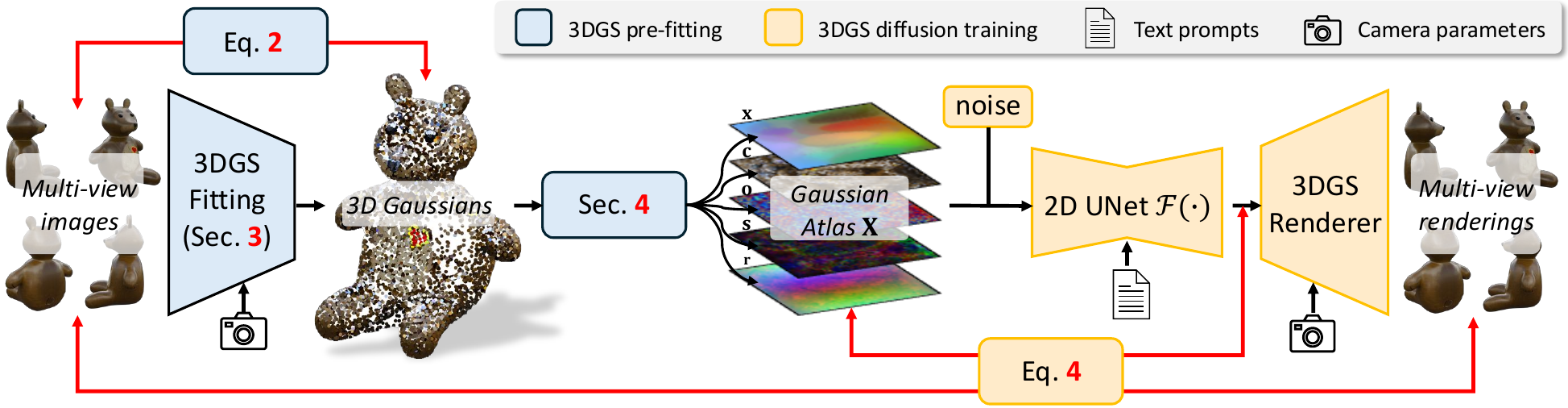}
    \vspace{-0.5em}
    \caption{\textbf{Repurposing 2D diffusion models for 3D Gaussian generation.} Our pipeline consists of two stages. In the 3DGS pre-fitting stage (\sectionautorefname~\ref{sec:dataset}), we pre-fit high quality 3D Gaussians for a diverse array of 3D objects with multi-view observations. The large-scale 3DGS fittings are then re-structured as Gaussian Atlas (\sectionautorefname~\ref{sec:methods}) with 2D representations for each 3DGS attribute. In the diffusion model training stage (\sectionautorefname~\ref{sec:diffusion}), we leverage the transformed 2D Gaussian atlases to repurpose a pretrained latent diffusion model (the 2D UNet denoiser $\mathcal{F}$) and ultimately achieve 3D content generation.}
    \label{fig:method}
    \vspace{-0.5em}
\end{figure*}

\section{Formulating 3D Gaussians as 2D Atlas} \label{sec:methods}
% \subsection{A 2D Representation of 3D Gaussian}
In this section, we introduce a novel approach that transforms unorganized Gaussians in the 3D space to a dense 2D representation, namely \methodname, making it possible to repurpose 2D diffusion models, for example, Latent Diffusion (LD) \cite{Rombach_2022_CVPR}, for 3D generation tasks. 

%We first review fundamentals of the text-to-image generation model, Latent Diffusion \cite{Rombach_2022_CVPR}, and then explain how it can help generate 3D contents.

%Subsequently, we present our method that transforms unorganized 3D Gaussians into structured 2D grids to be used for finetuning 2D diffusion models. 

\vspace{0.5em}\noindent\textbf{Motivation.} LD can understand complex natural language and generate coherent 2D images, benefiting from the vast availability of over billions of paired text-image data \cite{schuhmann2022laion}. However, text-to-3D generation presents greater challenges due to two key reasons: \emph{(i)} the scarcity of large-scale datasets with 3D models comparable to those in 2D, as creating and annotating high-quality, textured 3D models remains both resource-intensive and time-consuming \cite{luo2023scalable}; and \emph{(ii)} the inherently higher-dimensional nature of 3D object representations, which impose complex geometric constraints, making it more difficult for diffusion models to interpret. These disparities in data availability and representational complexity between 2D and 3D motivate our approach to leveraging learned priors from pre-trained 2D diffusion models for 3D Gaussian generation. 

However, unstructured Gaussians in 3D space cannot be directly passed to 2D models, which require inputs $\mathbf{X}$ to have: \emph{(i)} only 2 spatial dimensions; \emph{(ii)} valid ``pixels'' at each vertex of a dense 2D grid; and \emph{(iii)} values within a specific distribution, either $\mathbf{X}\in[-1,1]$ for the VAE or $\mathbf{X}\sim\mathcal{N}(0, 1)$ for the denoiser. To make 3D Gaussians compatible with 2D diffusion models, we propose \textbf{\methodname}, a 2D representation of 3D Gaussians.

One simple idea of 2D transformation is to project 3DGS onto an image plane with their 3D coordinates and given camera parameters. However, this naive approach loses depth information entirely and disrupts the original 3D structure’s topology, which is essential for 3D generation as it relies on accurate representation of 3D continuity. This highlights the need for a method that not only maps 3D Gaussians onto a 2D plane but also preserves 3D continuity to some extent. Given these requirements, we argue that methods similar to UV texture unwrapping, which unfold the surface of a 3D geometry onto a 2D plane, are more suitable. However, UV maps are usually not universally applicable since careful designs are needed for different 3D geometries \cite{bai2023ffhq, guler2018densepose}. Therefore, instead of exact mappings between 3D geometries and their 2D UVs, we focus on one characteristic of UV unwrapping --- geometry flattening.

Specifically, our goal is to find a mapping function $\mathcal{M}(\cdot)$ that transforms the 3D positions $\{\mathbf{x}\in\mathbb{R}^3\}$ of 3D Gaussians to 2D planar coordinates $\{\mathbf{\hat{x}}\in\mathbb{R}^2\}$: $\mathcal{M}(\{\mathbf{x}\})\rightarrow \{\mathbf{\hat{x}}\}$. A natural approach is to parameterize such a function $\mathcal{M}$ as neural networks which can be optimized toward 2D and 3D consistency \cite{zhang2024flatten}. However, this approach requires repeated training of $\mathcal{M}$ on different objects which is not only time consuming but also makes the mapping process inconsistent between objects given that the heuristics for 2D flattening may be different for different geometries. As a result, diffusion models are not able to capture the irregular patterns and fail to generate meaningful contents. Experiments are provided in \sectionautorefname~\ref{sec:ablation} to support this claim. To this end, we seek a more coherent transformation between 3D and 2D with a simple and deterministic mapping function. 

We formulate our transformation process into three sequential stages, as outlined in \figureautorefname~\ref{fig:atlas}. In the first stage, we hypothesize a unit sphere $\mathcal{S}$, which is centered at the origin of 3D-axis with unit radius, to be parameterized by $N$ 3D points $\{s_i \in \mathbb{R}^3\}$ that are uniformly distributed on its surface. Sphere is a well-studied structure that offers multiple standards for 2D projection, which makes it suitable for being considered in the flattening process. We then translate the unorganized 3D Gaussians to the surface points $\{s_i \in \mathbb{R}^3\}$ of $\mathcal{S}$ using Optimal Transport (OT) \cite{burkard1999linear}. We call this process \emph{sphere offsetting}. Our method differs from~\cite{zhang2024gaussiancube} by offsetting 3D Gaussians directly onto the \emph{surface}, rather than vertices \emph{within} a volume. Our transportation process is also much faster due to the adaptive number of Gaussians.

After positioning the 3D Gaussians on the surface of a unit sphere, we employ equirectangular projection \cite{enwiki:1243350398} as our $\mathcal{M}$ to obtain flattened 2D coordinates $\{p_i \in \mathbb{R}^2\}$ of the 3D Gaussians. Lastly, in the \emph{plane offsetting} stage, we apply another OT to map $\{p_i\}$ to the vertices $\{q_i \in \mathbb{R}^2\}$ of a square 2D grid with spatial size $\sqrt{N}\times\sqrt{N}$, further reducing sparsity. In particular, since now $\mathcal{M}$ is a deterministic function, the mapping between $\{p_i\}$ and $\{q_i\}$ remains identical for all objects. This consistency allows us to perform OT in the last stage only once and reuse the computed indices for all objects. We refer to the final grid-like 2D representation of 3D Gaussians as \methodname, with each atlas $\mathbf{X}$ in the shape of $\sqrt{N}\times \sqrt{N}\times (||\mathbf{x}-\mathbf{s}|| + ||\mathbf{c}|| + ||\mathbf{o}|| + ||\mathbf{s}|| + ||\mathbf{r}||)$ entailing all attributes of 3D Gaussians.

\section{2D Diffusion for 3D Gaussian Generation} \label{sec:diffusion}

After transforming fitted 3D Gaussians onto 2D planes, we are able to fine-tune the pre-trained Latent Diffusion (LD) \cite{Rombach_2022_CVPR} with Gaussian atlases. 

\vspace{0.5em}\noindent\textbf{Preliminaries: Latent Diffusion.} LD is a family of models that generate 2D images based on text prompts. The core components of LD include a Variational AutoEncoder (VAE) and a UNet $\mathcal{F}(\cdot)$. The VAE encoder compresses images into lower-dimensional latents $l$ to facilitate efficient diffusion in the latent space. By injecting Gaussian noise to the latents, $\mathcal{F}$ can be trained through self-supervised denoising via v-parameterization \cite{salimans2022progressive}:
% \begin{equation}
%     \mathcal{L}_{\text{diffusion}} = \mathbb{E}_{\mathbf{x}_0, \mathbf{z}, t} \left[ \left\| \mathbf{z} - \mathbf{z}_\theta (\mathbf{x}_t, t) \right\|^2 \right],
% \end{equation}
\begin{equation} \label{eq:diff}
    \mathcal{L}_{diff}=\mathbb{E}_{l_0, \mathbf{z}, t} \left[ \left\| \nabla_{l_t} \mathbf{z} - \nabla_{l_t} \mathcal{F}(l_t, t) \right\|^2 \right],
\end{equation}
where $l_t$ is the noisy latent at timestamp $t$, $\nabla_{l_t} \mathbf{z}$ is the `velocity' added to $l_t$. During inference, a sample can be generated through the reverse diffusion process by iteratively forwarding $\mathcal{F}$ for denoising. The VAE decoder then upsamples the generated latent back to the original RGB space.

\vspace{0.5em}\noindent\textbf{Finetuning LD with Gaussian Atlases.} The standard fine-tuning approach for LDs involves VAE-based encoding and decoding \cite{ke2024repurposing}. However, in the supplementary materials, we argue that such auto-encoding is unsuitable for Gaussian atlases - the Gaussian attributes differ from the natural images used to train the LD. Finetuning LD UNet with 2D atlases is feasible only if the distributions of the atlases are aligned with the VAE-encoded latents. We therefore standardize the 2D atlases using the pixel-wise mean and standard deviation computed from the entire \datasetname. This normalization prevents extreme high- or low-frequency pixel values while preserving the characteristics of the original VAE-encoded image latents, ultimately accelerating the fine-tuning process.

% 3D Gaussians are associated with five different attributes. The 3D location (mean) $\mathbf{x}$, albedo $\mathbf{c}$, and scale $\mathbf{s}$ are three-channel features, while opacity $\mathbf{o}$ is single-channel, and rotation $\mathbf{r}$ is typically represented as four-channel quaternions. Since the 2D UNet from LD was originally trained on four-channel latents, we pad each of the three-channel attributes with the one-channel opacity and repeat the input layer of the UNet four times \cite{ke2024repurposing, fu2025geowizard} to accommodate the combination of all attributes as inputs. The final Gaussian atlases, which are passed to the UNet for fine-tuning, have a shape of $128 \times 128 \times 16$. We also normalize the Gaussian atlases into standard Gaussian distributions, using pixel-wise means and standard deviations calculated from all the 3DGS fittings in \datasetname. When rendering, we unpack different attributes accordingly and average out the three repetitions of opacity.

\begin{figure*}[t]
    \centering
    \includegraphics[width=0.95\linewidth]{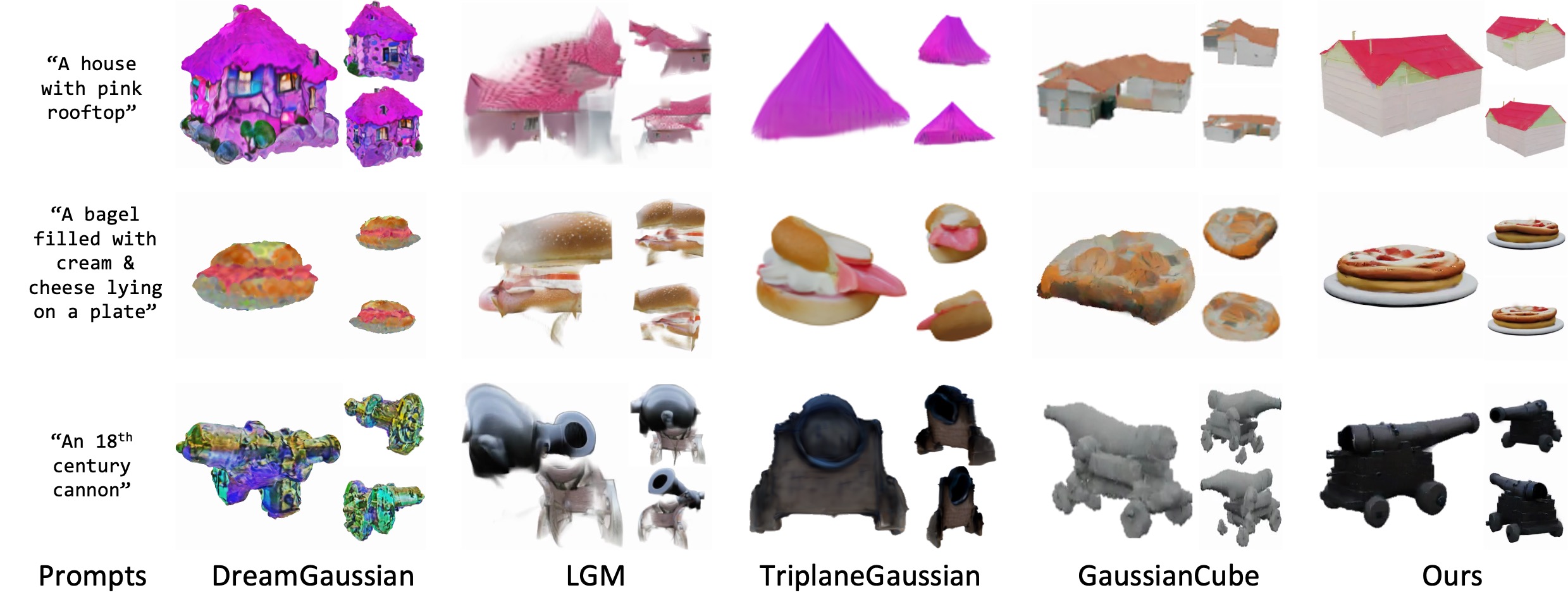}
    \vspace{-0.5em}
    \caption{\textbf{Qualitative Comparisons.} Our 3D generations exhibit the highest quality, minimal artifacts, and the best alignment with text prompts. In contrast, DreamGaussian~\cite{tang2023dreamgaussian}, LGM~\cite{tang2025lgm}, and TriplaneGaussian~\cite{zou2024triplane} struggle to produce natural and complete 3D assets, while GaussianCube~\cite{zhang2024gaussiancube} fails to capture key phrases in text prompts, such as \emph{``on a plate"} and \emph{``pink rooftop"}.}
    \vspace{-0.5em}
    \label{fig:comp}
\end{figure*}

3D Gaussians are characterized by five attributes: the 3D location (mean) $\mathbf{x}$, albedo $\mathbf{c}$, and scale $\mathbf{s}$, each represented as three-channel features; opacity $\mathbf{o}$, which is a single-channel feature; and rotation $\mathbf{r}$, typically represented as a four-channel quaternion. Since the LD UNet was originally trained on four-channel latents, we pad each three-channel attribute with the one-channel opacity and repeat the input layer of the UNet four times \cite{ke2024repurposing, fu2025geowizard} to accommodate all attributes as inputs. The final Gaussian atlases, used for fine-tuning the UNet, have a shape of $128 \times 128 \times 16$. During rendering, we appropriately unpack the different attributes and average the three repetitions of opacity.

We finetune the LD UNet $\mathcal{F}$ with a combination of diffusion loss (\equationautorefname~\ref{eq:diff}) and photometric loss between the renderings from denoised Gaussians and the reference images. The final objective for tuning $\mathcal{F}$ becomes:
\begin{equation}
    \lambda_{diff}\mathcal{L}_{diff} + \lambda_{rgb}\mathcal{L}_{rgb} + \lambda_{mask}\mathcal{L}_{mask} + \lambda_{lpips}\mathcal{L}_{lpips},
\end{equation}
where $\mathcal{L}_{rgb}, \mathcal{L}_{mask}$ are L1 loss on RGB renderings and accumulated opacity maps and $\mathcal{L}_{lpips}$ is the perceptual loss. $\lambda$s are the loss weights. Our pipeline is outlined in \figureautorefname~\ref{fig:method}.

%% file: sec/5_experiment.tex
\section{Experiments}
In this section, we benchmark and evaluate our proposed methods on zero-shot text-to-3D generation, one of the most fundamental tasks in generative 3D modeling.

\begin{figure*}
    \centering
    \includegraphics[width=0.95\linewidth]{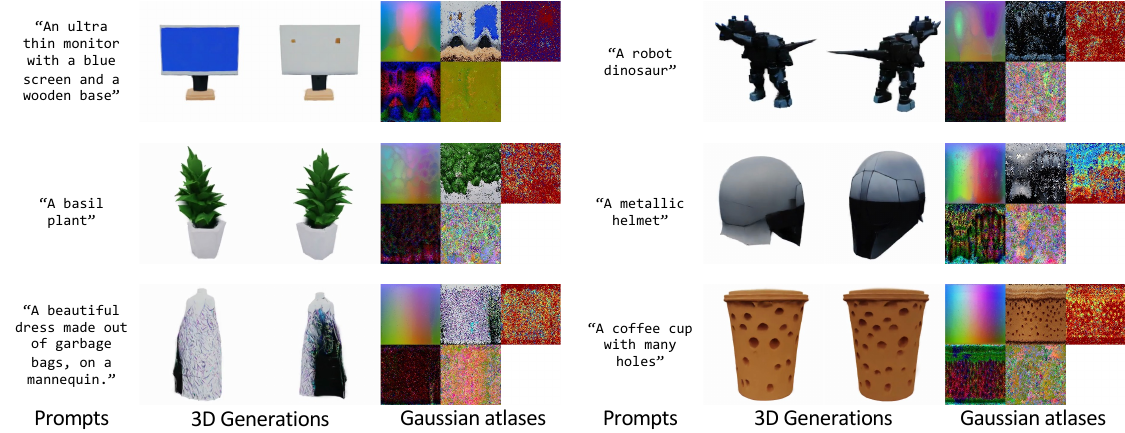}
    \vspace{-0.5em}
    \caption{\textbf{Additional qualitative results.} Our method effectively repurposes 2D diffusion models for high-quality 3D contents. The generated Gaussian atlases are presented in the order from top left to bottom right: 3D location $\mathbf{x}$, albedo $\mathbf{c}$, color-coded opacity $\mathbf{o}$, normalized scale $\mathbf{s}$, and the last three channels of normalized quaternion $\mathbf{r}$. \emph{More results can be found in the supplementary materials.}}
    \vspace{-0.5em}
    \label{fig:results}
\end{figure*}

\vspace{0.5em}\noindent\textbf{Implementation Details.} Given the upper bound $\tau = 36,864$ of valid 3D Gaussians during the pre-fitting stage, we set the total number of 3D Gaussians per Gaussian atlas to $N = 16,384 = 128 \times 128$. This choice of $N$ accommodates the large variance of 3D Gaussians in \datasetname.
%(\tableautorefname~\ref{tab:3dgs_data}, \figureautorefname~\ref{fig:GO} (d)).

% A prefix `\texttt{Gaussian Atlas showing }' was appended at the front of each prompt to distinguish Gaussian atlases and natural images.

We initialize the 2D UNet $\mathcal{F}$ from a LD pretrain checkpoint \cite{schuhmann2022laion} and use CLIP \cite{radford2021learning} to encode text prompts. The UNet is then fine-tuned with classifier-free guidance by randomly zeroing out the text encodings with a probability of 20\%. We employ the AdamW optimizer \cite{loshchilov2017decoupled} with a learning rate of $5 \times 10^{-5}$ and a batch size of 64. Exponential Moving Average (EMA) and mixed precision training are enabled for stable and efficient training. For fair comparisons with the baseline method~\cite{zhang2024gaussiancube},  only approximately 130K atlases from \datasetname\ were used for training. The paired text prompts were from \cite{luo2023scalable} and the multi-view references were from \cite{zuo2024high}. The LD UNet was finetuned for 1M steps on 8 A100 GPUs. 

For inference, we start from 2D random noise in the same shape as a Gaussian atlas and generate clean samples via reverse diffusion using the adaptive DPMsolver$++$ \cite{lu2022dpm} with a guidance scale of 3.5. A 3DGS sample can be generated and rendered in less than 5 seconds.

\vspace{0.5em}\noindent\textbf{Metrics.} Reference based metrics are not feasible for evaluating zero-shot text-to-3D generations. Following \cite{poole2022dreamfusion, jain2022zero, zhang2024gaussiancube},  we evaluate the alignment of text prompts and 10 random 2D renderings of the 3D generations with CLIP score \cite{radford2021learning} as well as VQA score \cite{lin2024evaluating} on 100 diverse test prompts. Additional user studies and qualitative comparisons were also conducted for comprehensive evaluations.

% , containing a variety of captions for simple objects (e.g., `\texttt{A blue car}'), counterfactual objects (e.g., `\texttt{A metal apple with gold stem}'), and complex objects (e.g., `\texttt{A house with colorful rooftop}').

\vspace{0.5em}\noindent\textbf{Comparisons.} We compare our proposed method against four representative methods in the field of 3D Gaussian generation: An optimization based method --- DreamGaussian \cite{tang2023dreamgaussian}; A method uses also a 2D model --- LGM \cite{tang2025lgm}; A method generates also 2D representations --- TriplaneGaussian \cite{zou2024triplane}; and the state-of-the-art 3D Gaussian generator with a 3D diffusion model --- GaussianCube \cite{zhang2024gaussiancube}. For LGM and TriplaneGaussian, we utilize MVDream~\cite{shi2023mvdream} to generate 2D images from text prompts to initiate 3D generations. Notably, since there are no open-source implementations of the relevant works that also adopt 2D representations of 3D content \cite{liu2024pi3d, mercier2024hexagen3d, elizarov2024geometry}, direct comparisons are not possible.

\begin{table}[t]
    \centering
    \resizebox{\linewidth}{!}{%
    \begin{tabular}{c|ccc}
    \toprule
       Methods  & CLIP score $\uparrow$  & VQA score $\uparrow$ & \# Gaussians $\downarrow$\\
       \hline
       DreamGaussian \cite{tang2023dreamgaussian} & 20.52 & 0.37 & 40K\\
       LGM \cite{tang2025lgm} & 20.28 & 0.35 &66K \\
       TriplaneGaussian \cite{zou2024triplane} & 21.10 & 0.46& \cellcolor{best_color}{\textbf{16K}}\\
       GaussianCube \cite{zhang2024gaussiancube} & 22.31 & 0.52 &33K\\
       GaussianAtlas (Ours) & \cellcolor{best_color}{\textbf{23.20}} & \cellcolor{best_color}{\textbf{0.61}} & \cellcolor{best_color}{\textbf{16K}}\\
       \bottomrule
    \end{tabular}
    }
    % \caption{Qualitative comparisons on zero-shot text-to-3D generation. With $\sim\frac{1}{4}\times$ and $\sim\frac{1}{2}\times$ number of Gaussians, our method achieves on par performances to the state of the arts in terms of CLIP similarity scores.}
    \vspace{-0.5em}
    \caption{\textbf{Qualitative comparisons.} Our method achieves performance comparable to the state-of-the-art in terms of CLIP similarity scores, with the minimum number of 3D Gaussians.}
    \vspace{-0.5em}
    \label{tab:metrics}
%     # 0.613179 0.23202 # ours
% # 0.517087 0.22312299999999996 # GaussianCube
% # 0.46361900000000006 0.211032 # TriplaneGaussian
\end{table}

\subsection{Results}

\noindent\textbf{Qualitative results.} We present text-to-3D generation results in \figureautorefname~\ref{fig:comp}. Compared to DreamGaussian and LGM, our generations exhibit high visual fidelity without excessive details. In contrast to TriplaneGaussian and GaussianCube, our method aligns closely with the text prompts, reflecting more precise prompt conditioning. According to the additional results shown in \figureautorefname~\ref{fig:results}, our fine-tuned 2D diffusion model is capable of generating high-quality Gaussian atlases. Consequently, our generated 3D contents are coherent and self-contained, with no seams or artifacts that would typically be observed from UV unwrapping methods. 
%\emph{More qualitative results are in the supplementary materials.}

\vspace{0.5em}\noindent\textbf{Quantitative results.} In \tableautorefname~\ref{tab:metrics}, we present the CLIP and VQA scores for comparison methods. Our method outperforms all counterparts while generating the minimum number of Gaussians required. Specifically, it uses only \textbf{$\frac{1}{2}$} the number of Gaussians and the number of training steps on 3D data compared to GaussianCube, achieving a CLIP score higher by 0.9 and a VQA score higher by 17\%. This further supports our claim that pretrained 2D diffusion models can be repurposed for 3D content generation.

\begin{figure}[t]
      \centering
      \includegraphics[width=0.99\linewidth]{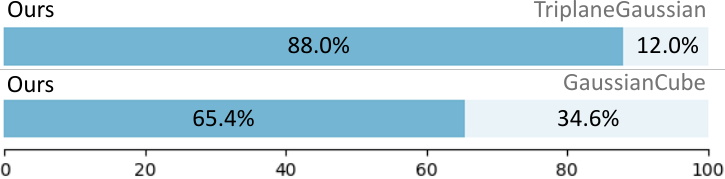}
      \vspace{-0.5em}
      \caption{\textbf{User study results.} Our method outperforms state-of-the-art methods \cite{zhang2024gaussiancube,zou2024triplane} in user preferences regarding generation quality and alignment with text prompts.}
      \vspace{-0.5em}
      % \caption{Optimization-based flattening approach yields no consistent patterns and leads to finetuning difficulty. We show two 3D objects (a) ane their corresponding Gaussian Atlas (3D location $\mathbf{x}$ only) obtained from our proposed flattening approach (b) and the optimization-based approach (c). After training to the same steps, only noisy Gaussians (d) and atlases (e) are generated from the prompt `\texttt{A light blue chair}'.}
      \label{fig:user_study}
  \end{figure}

\vspace{0.5em}\noindent\textbf{User studies\footnote{The user study was conducted entirely by researchers at Stanford.}.} The user study was conducted with diverse participants who were asked to assess the overall generation quality to given text prompts. They performed pairwise comparisons between generations by our method and those from a competing method. We calculated the win rate from $>2,500$ valid responses. As shown in \figureautorefname~\ref{fig:user_study}, over 65\% of users preferred the 3D content generated by our method when compared to GaussianCube. The preference rate was even more pronounced against TriplaneGaussian, with 88\% of users favoring our method.

\subsection{Discussions and Ablations} \label{sec:ablation}
This work takes a step toward unifying 2D and 3D generation by transforming 3D Gaussians onto 2D planes, allowing the direct fine-tuning of a pre-trained 2D diffusion model for generating 3D contents. In this section, we conduct extensive studies and ablative experiments to validate our proposed approach from three key perspectives.

\begin{figure}
      \centering
      \includegraphics[width=0.99\linewidth]{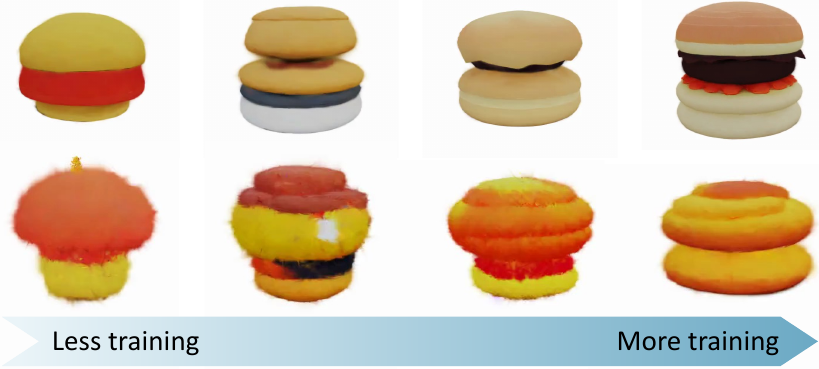}

      \vspace{0.5em}
    \resizebox{\linewidth}{!}{%
    \begin{tabular}{c|ccc}
    \toprule
       Methods & \# Training steps & CLIP score $\uparrow$   &VQA score $\uparrow$\\
       \hline
       Train from scratch & 500K & 19.33 & 0.23\\
       Repurpose 2D LD & 500K
       &  \cellcolor{best_color}{21.61} & \cellcolor{best_color}{0.49} \\
        \hline
       Train from scratch & 1M & 20.85 & 0.40 \\
       Repurpose 2D LD & 1M
       & \cellcolor{best_color}{\textbf{23.20}} & \cellcolor{best_color}{\textbf{0.61}} \\
       \bottomrule
    \end{tabular}
    }
    \vspace{-0.5em}
    \caption{Finetuning from a pretrained 2D diffusion model leads to \emph{faster} generalization. \textbf{Top:} 3D generations at different training checkpoints from finetuning (top row) and training from scratch (bottom row) using the same prompt `\texttt{A toy hamburger with beef patty \& cheese}'. \textbf{Bottom:} Quantitative comparisons at checkpoints with different training steps.}
    \vspace{-0.5em}
      % \caption{Optimization-based flattening approach yields no consistent patterns and leads to finetuning difficulty. We show two 3D objects (a) ane their corresponding Gaussian Atlas (3D location $\mathbf{x}$ only) obtained from our proposed flattening approach (b) and the optimization-based approach (c). After training to the same steps, only noisy Gaussians (d) and atlases (e) are generated from the prompt `\texttt{A light blue chair}'.}
      \label{fig:scratch}
  \end{figure}

 \vspace{0.5em}\noindent$\bullet$\ \ \emph{Learned knowledge in text-to-image diffusion models is universally transferable.} Without pre-training on large-scale 2D data, the model struggles to develop a comprehensive understanding of natural language and content generation when trained solely on 3D data. As shown in \figureautorefname~\ref{fig:scratch}, we effectively repurpose the pre-trained 2D diffusion model with faster convergence and higher-quality 3D generation. The qualitative results show that, compared to training from scratch, our approach produces more coherent and structured 3D assets earlier in training. The quantitative results further support this finding, as our repurposed models consistently achieve higher scores in different training steps.

\begin{figure}[t]
      \centering
      \includegraphics[width=0.99\linewidth]{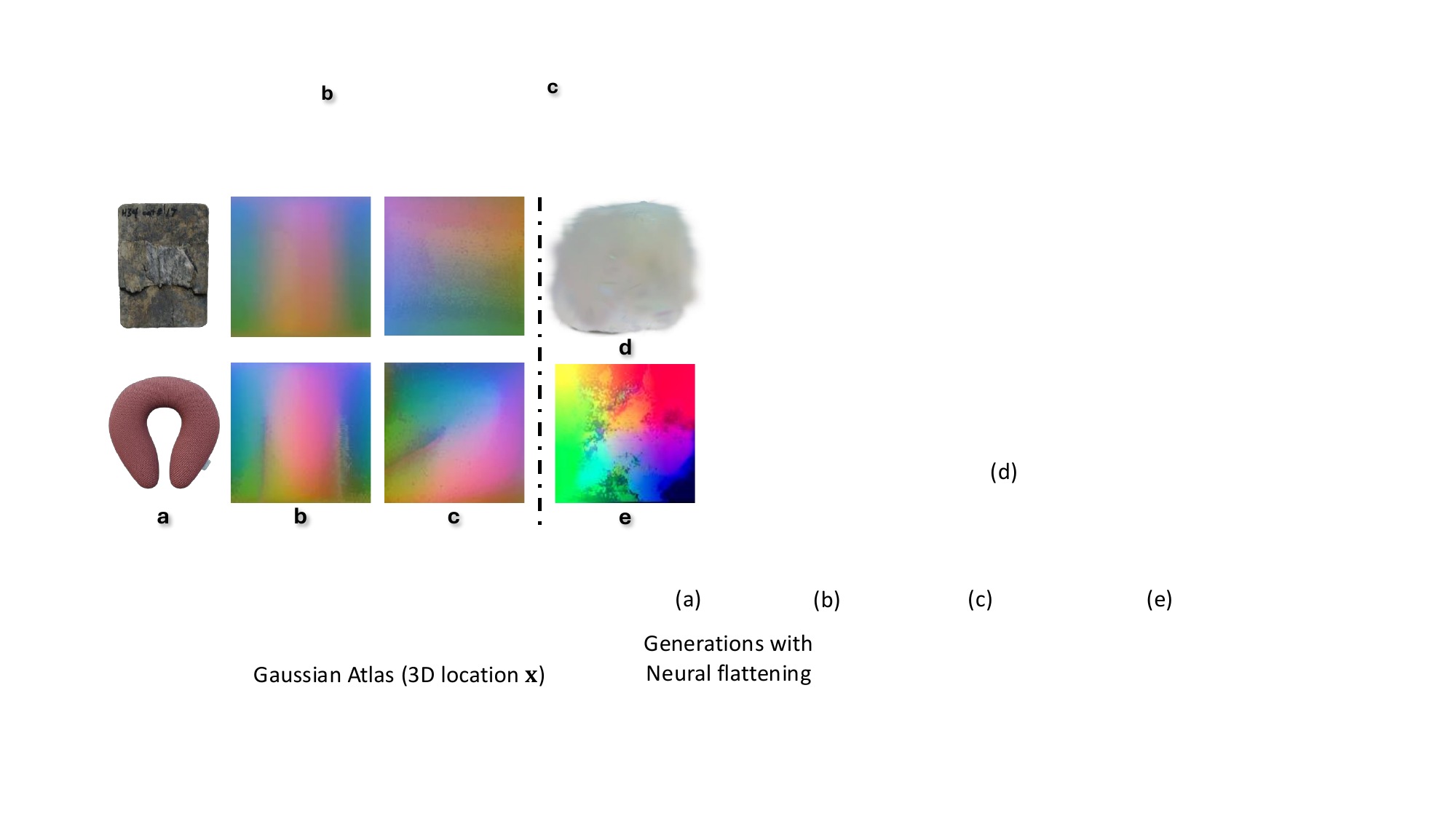}
      \vspace{-0.5em}
      \caption{The optimization-based flattening approach results in inconsistent patterns that are hard for fine-tuning. We show two objects (a) and their Gaussian atlases (3D location $\mathbf{x}$ only) obtained using our proposed flattening approach (b) and the optimization-based approach~\cite{zhang2024flatten} (c), which generates only noisy Gaussians (d) and atlases (e) after training for the same number of steps.}
      \vspace{-0.5em}
      % \caption{Optimization-based flattening approach yields no consistent patterns and leads to finetuning difficulty. We show two 3D objects (a) ane their corresponding Gaussian Atlas (3D location $\mathbf{x}$ only) obtained from our proposed flattening approach (b) and the optimization-based approach (c). After training to the same steps, only noisy Gaussians (d) and atlases (e) are generated from the prompt `\texttt{A light blue chair}'.}
      \label{fig:flatten}
  \end{figure}

\vspace{0.5em}\noindent$\bullet$\ \ \emph{Gaussian atlases yield consistent and capturable visual patterns, which makes the pre-trained 2D diffusion model easier to generalize towards.} To validate the claim, we ablate on the ``flattening'' strategy and adopted an optimization based approach \cite{zhang2024flatten} to find the most natural cut for each object and perform 3D-to-2D transformation via a combination of neural networks. We train the networks on a per-object basis for the entire \datasetname, which roughly took another A100 GPU year. The flattened 3D Gaussians are then used to finetune the LD UNet by following the same training procedure. We visualize generation results in \figureautorefname~\ref{fig:flatten}. With capturable visual patterns, our proposed flattening strategy leads to faster model convergence and much higher generation quality at the same training step.

 % 100/510863

\begin{figure}[h]
      \centering
      \includegraphics[width=0.99\linewidth]{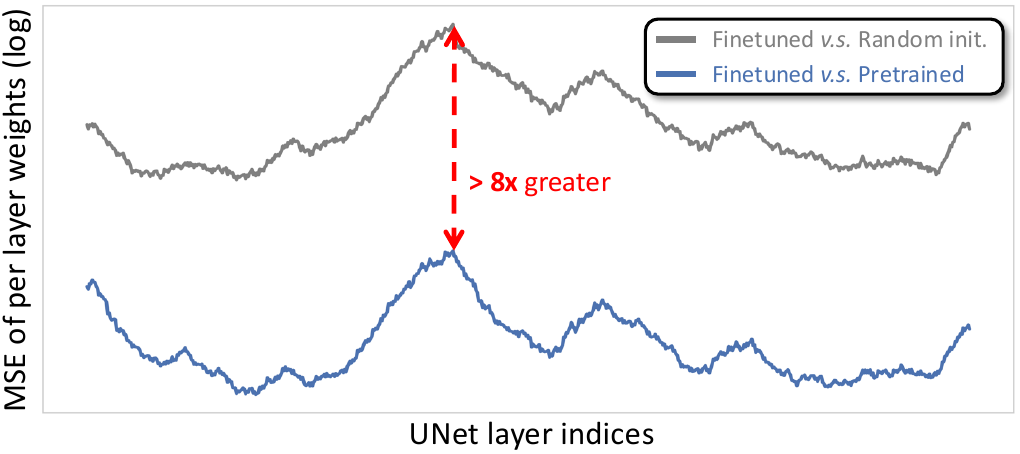}
      \vspace{-0.5em}
      \caption{Log scale mean squared errors (smoothed) of per layer weights between different UNets. Finetuning from a pre-trained LD UNet leads to smaller weight differences.}
      \vspace{-0.5em}
      \label{fig:weights}
  \end{figure}

\vspace{0.5em}\noindent$\bullet$\ \ \emph{A pre-trained 2D diffusion model serves as a better initialization for 3D generation.} In \figureautorefname~\ref{fig:weights}, we plot the weight differences between our fine-tuned UNet and both a randomly initialized UNet and the pre-trained UNet directly from LD. We observe overall small changes in the weights from the pre-trained UNet. Even at the layer with the greatest change, the difference is still $8\times$ smaller than that from a random initialization. This indicates that the pre-trained LD weights provide better starting points for 3D generation, facilitating easier convergence to a local minimum.

%% file: sec/6_conclusion.tex
\section{Conclusion}
This paper presents an advancement in unifying 2D and 3D content generation by repurposing a pre-trained 2D diffusion models to generate 3D Gaussians. Historically, fine-tuning 2D diffusion models with 3D Gaussians was infeasible. We introduce \methodname, a novel approach that transforms unorganized 3D Gaussians into a structured 2D grid, thereby enabling the effective utilization of 2D networks for 3D generation. We also present \datasetname, a large-scale dataset comprising 205,737 3D Gaussian fittings of diverse 3D objects, which facilitates diffusion model training. Experiments demonstrate that our method achieves state-of-the-art 3D generation results using significantly fewer Gaussians. We also provide extensive discussions and ablation studies that validate our claims and offer valuable insights into the effectiveness of our approach.

\vspace{-7pt}
\paragraph{Acknowledgments.} Tiange Xiang, Scott Delp, Ehsan Adeli, and Li Fei-Fei's work at Stanford were partially funded by the NIH Grant R01AG089169 and P41EB027060, Panasonic Holdings Corporation, the Gordon and Betty Moore Foundation, the Jaswa Innovator Award, Stanford HAI, Stanford HAI graduate fellowship, and Stanford Wu Tsai Human Performance Alliance. We thank Fei Jiang and Moustafa Meshry for their insightful feedback on this work.

%% file: sec/X_suppl.tex
\clearpage
\setcounter{page}{1}
\maketitlesupplementary
\appendix

\section{Supplementary for \datasetname}

\begin{table*}[th]
    \centering
    \begin{tabular}{c|c|ccc}
    \toprule
        Methods & Venue &\# of 3DGS fittings & \# of Gaussians per fitting \\
        \hline
        GSD \cite{mu2024gsd} & ECCV 2024 & 6,000 & 1,024 \\
        
        GVGen \cite{he2024gvgen} & ECCV 2024 & $\sim$46,000 & 32,768 \\
        GaussianCube \cite{zhang2024gaussiancube} & NeurIPS 2024 & 125,653 & 32,768 \\
        ShapeSplat \cite{ma2024shapesplat} & 3DV 2025 & $\sim$65,000 & $>$20,000 \\
        \textbf{\datasetname\ (ours)} & - & 205,737 & 10,435 \\
        \bottomrule
    \end{tabular}
    \vspace{-0.5em}
    \caption{\datasetname\ contains large-scale 3DGS fittings with adaptive number of Gaussians, which allows various applications.}
    \label{tab:dataset_compare}
\end{table*}

\subsection{Fitting Details}

As described in Section~\ref{sec:dataset}, we adopted Scaffold-GS \cite{lu2024scaffold} as the base model for per-object 3D Gaussian fitting. We observed duplicated Gaussians when the number of offsets was large, even at the default value. Therefore, we reduced the number to 4 to allow more anchors to be initialized with random values. The 3D Gaussians were then optimized according to the objective in \equationautorefname~\ref{eq:3dgs_fit}, with $\lambda_{rgb}'$ set to 0.8, $\lambda_{ssim}'$ set to 0.2, $\lambda_{lpips}'$ set to 0.02, and $\lambda_{reg}'$ set to 0.01.

\subsection{3DGS Quality of \datasetname}

\begin{figure}[t]
    \centering
    \includegraphics[width=0.99\linewidth]{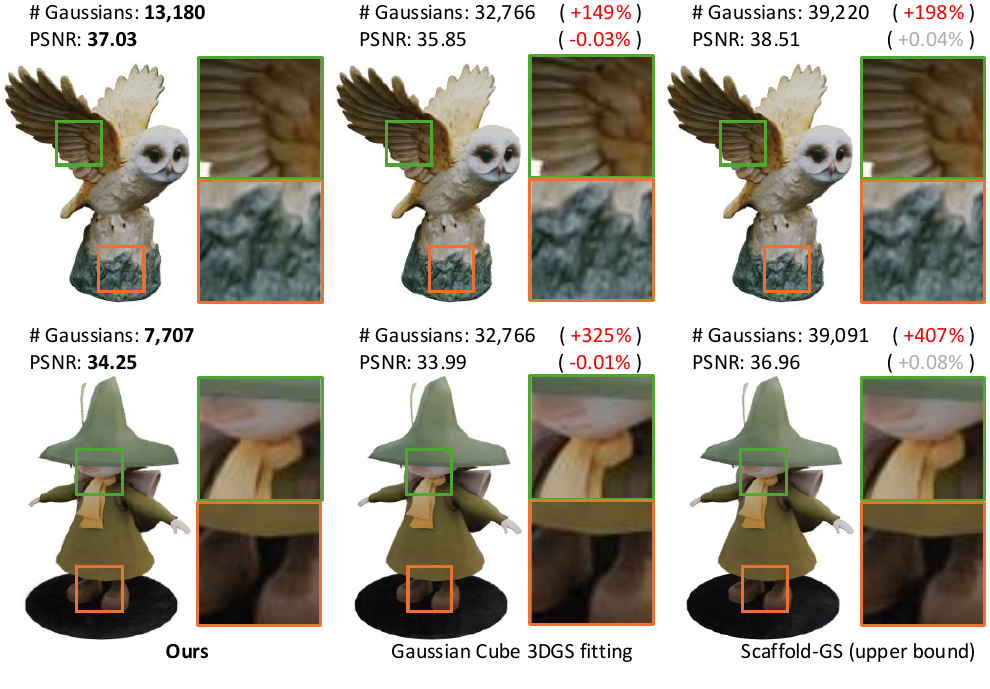}
    \vspace{-0.5em}
    \caption{\textbf{Comparisons of 3DGS fitting methods.} Our method achieves high fitting quality comparable to the upper bound with significantly fewer valid 3D Gaussians with positive opacity.}
    \vspace{-0.5em}
    \label{fig:dataset_comp}
\end{figure}

We compare our proposed 3DGS fitting method in Section~\ref{sec:dataset} with the densification-constrained fitting method introduced by GaussianCube \cite{zhang2024gaussiancube} and the upper-bound method, Scaffold-GS \cite{lu2024scaffold}, in \figureautorefname~\ref{fig:dataset_comp}. Compared to GaussianCube, our method achieves better rendering quality. Compared to the upper bound, our results show no significant visual degradation while using significantly fewer 3D Gaussians. As shown in \figureautorefname~\ref{fig:dataset_comp}, the rendering quality of fitted 3DGS does not improve with an increased number of Gaussians. This observation confirms that many Gaussians contribute minimally to the final rendering quality and can be merged or pruned during fitting.

%\emph{Our method achieves high fitting quality comparable to the upper bound with significantly fewer valid 3D Gaussians (positive opacity)}.

\subsection{Comparison with Other 3D Gaussian Datasets}

\datasetname\ is a large-scale dataset consisting of high-quality 3D Gaussian fittings for a wide range of objects. We note that there are a few other studies that also fit per-object 3D Gaussians to assist in the training of diffusion models. A direct comparison is provided in \tableautorefname~\ref{tab:dataset_compare}.

% \subsection{Fitting Details}

% As mentioned in \ref{sec:dataset}, we adopted the Scaffold-GS \cite{lu2024scaffold} as the base model for per-object 3DGS fitting. We observe duplicated Gaussians when the number of offset is large, even at the default value. We therefore lower the number to be 4 to allow more anchors to be initialized at random values. 3D Gaussians are then optimized towards the objective in \equationautorefname~\ref{eq:3dgs_fit} with $\lambda_{rgb}'$ to 0.8, $\lambda_{ssim}'$ to 0.2, $\lambda_{lpips}'$ to 0.02, and $\lambda_{reg}'$ to 0.01.

% \subsection{Comparison with Other 3D Gaussian Datasets}

% \datasetname\ is the first large-scale dataset consisting of high-quality 3D Gaussian fittings for a wide range of objects. We note that there are a few other studies that also fit per-object 3D Gaussians to assist the training of diffusion models. A direct comparison is provided in \tableautorefname~\ref{tab:dataset_compare}. We will release \datasetname\ to public to benefit research in relevant domains.

\section{Supplementary for \methodname\ Formulation}

\subsection{Formulation Details}

As discussed in Section \ref{sec:methods}, the adaptive nature of 3DGS fittings in \datasetname\ enables faster transformation of 3DGS to 2D \methodname\ compared to the similar process in \cite{zhang2024gaussiancube}. Notably, the computation time is significantly reduced during the non-square Optimal Transport step for \emph{sphere offsetting}, since the number of 3D positions $\{\mathbf{x}\}$ is typically much smaller than $N$, the number of surface points $\{s\}$ on the standard sphere $S$. After sphere offsetting, for 3DGS fittings that have a smaller size than $N$, we pad additional Gaussians by duplicating the ones with the smallest scales and set their opacity to 0. For 3DGS fittings that have more Gaussians than $N$, we prune the Gaussians with the smallest scales before sphere offsetting. In practice, constructing one Gaussian atlas takes an average of approximately one minute.

\section{Supplementary for Finetuning LD with \methodname} 

\subsection{Training Diffusion Model Without VAEs}

\begin{figure}[t]
    \centering
    \includegraphics[width=0.99\linewidth]{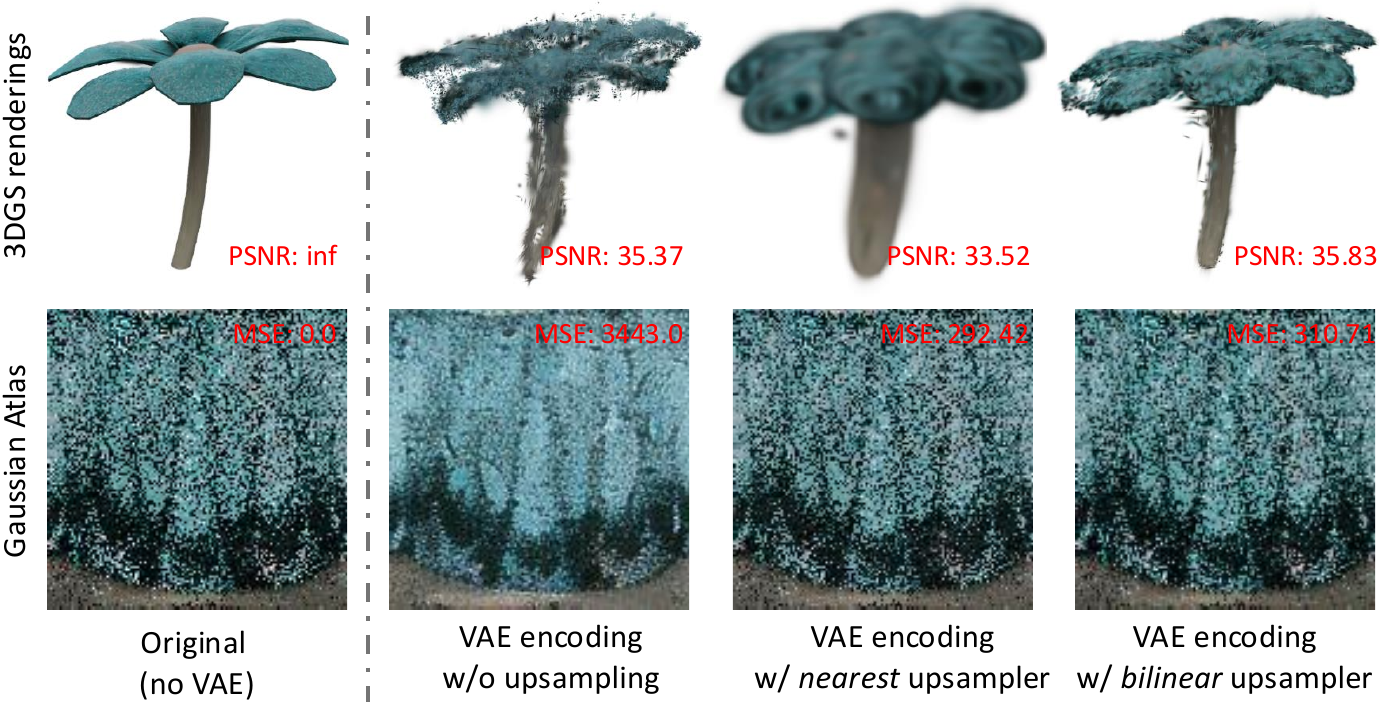}
    \vspace{-0.5em}
    \caption{\textbf{Latent diffusion VAE degrades 3D Gaussian quality.} In the bottom row, we present a transformed Gaussian atlas and its reconstruction with and without VAE auto-encoding. When the atlas is up-sampled to a higher resolution, no obvious differences are observed in the atlas. However, after rendering the decoded Gaussians to images, we can observe much higher disparities, resulting in significantly reduced PSNR and visual quality. This suggests that incorporating the VAE for 3DGS diffusion is impractical.}
    \label{fig:vae}
\end{figure}

The typical fine-tuning approach involves VAE encoding and decoding \cite{ke2024repurposing}. However, we argue that such a VAE auto-encoding is inappropriate for Gaussian atlases due to three reasons: \emph{(i)} VAE encoding is a lossy compression of the atlases, whose accuracy is crucial for Gaussian rendering. We demonstrate the impact of using VAE for auto-encoding Gaussian atlases in \figureautorefname~\ref{fig:vae}; \emph{(ii)} Even at the maximum bound $\tau=36,864=192\times192$, the number of Gaussians is significantly smaller than the number of pixels required for VAE input (e.g., $768\times768$). Naive up-sampling of the Gaussian Atlas $\mathbf{X}$ would significantly increase computational costs; \emph{(iii)} Even with scaling and shifting, $\mathbf{X}$\ does not visually resemble the natural RGB images originally used for VAE training, but rather analogizes latent features. Considering these, we remove the original VAE and fine-tune the LD UNet directly on Gaussian atlases. We make appropriate modifications to the input and output layers of the UNet to accommodate all Gaussian attributes with the correct number of channels \cite{ke2024repurposing}.

\subsection{More Details}

\paragraph{Diffusion model training.} We set $\lambda_{diff}$ to 1.0, $\lambda_{rgb}$ to 10.0, $\lambda_{mask}$ to 1.0, and $\lambda_{lpips}$ to 1.0 to finetune the LD UNet. At each training step, we randomly sample one view of 2D renderings to compute the photometric losses. We enable gradient checkpoint to save GPU memory, which allows a local batch size of 8 for each GPU, with a total batch size of 64 across 8 GPUs.

% \subsection{More Details}

% \paragraph{\methodname\ transformation is efficient.} As mentioned in \ref{sec:methods}, due to the adaptive nature of 3DGS fittings in \datasetname, transforming 3DGS to 2D \methodname\ can be faster than the similar process as in \cite{zhang2024gaussiancube}. Specifically, the computation time is mainly saved at the non-square Optimal Transport for \emph{sphere offsetting}, since the 3D positions $\{\mathbf{x}\}$ are usually much fewer than $N$ --- the number of surface points $\{s\}$ on the standard sphere $S$. In practice, constructing one Gaussian atlas takes an average of $\sim$60 seconds.

% \paragraph{Details on diffusion model training.} 
% We set $\lambda_{diff}$ to 1.0, $\lambda_{rgb}$ to 10.0, $\lambda_{mask}$ to 1.0, $\lambda_{lpips}$ to 1.0. At each training step, we randomly sample one view of 2D renderings to calculate the photometric losses. We enable gradient checkpoint to save GPU memories, which allows a local batch of 8 for each GPU, with a total batch size of 64 across 8 GPUs.

\section{More Results}

\subsection{More Comparisons}

\begin{figure}[t]
    \centering
    \includegraphics[width=0.99\linewidth]{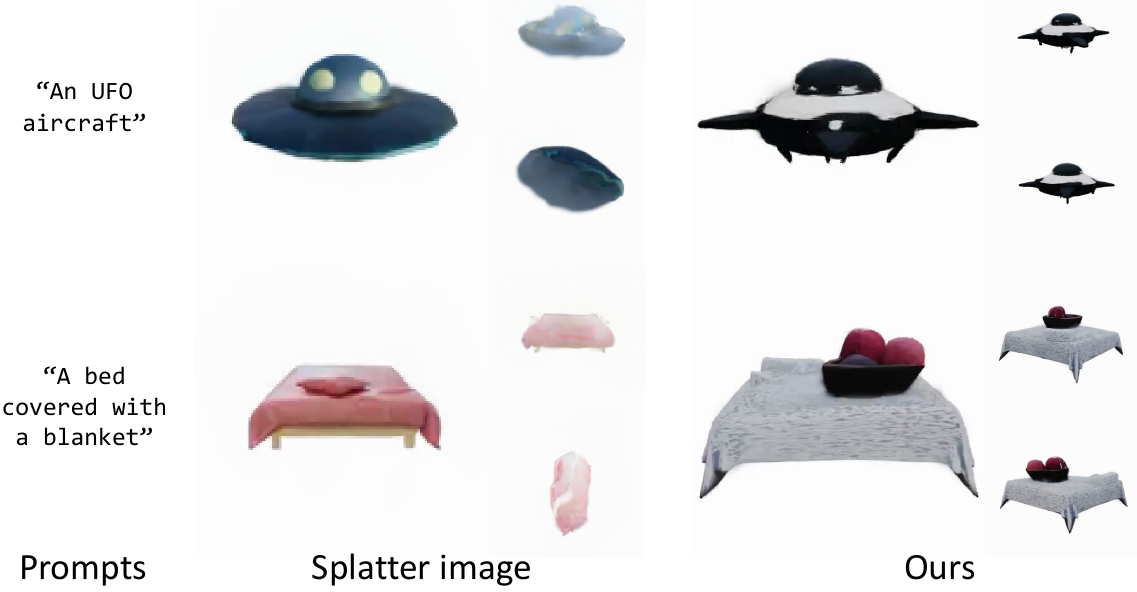}
    \vspace{-0.5em}
    \caption{\textbf{Comparison with a 2D generation approach~\cite{szymanowicz2024splatter}.}  Note that we provide MVDream~\cite{shi2023mvdream} generated 2D images (shown as the main images) to initialize 3D generations of \cite{szymanowicz2024splatter} (shown as the smaller images).}
    \vspace{-0.5em}
    \label{fig:appendx_comp_2d}
\end{figure}

Since Omages~\cite{yan2024object} does not support text-to-3D generation and the implementations of PI3D~\cite{liu2024pi3d}, HexaGen3D~\cite{mercier2024hexagen3d}, and GIMDiffusion~\cite{elizarov2024geometry} are not publicly available, we provide additional qualitative comparisons against Splatter Image~\cite{szymanowicz2024splatter}, a method that also generates 2D representations of 3D objects. As shown in \figureautorefname~\ref{fig:appendx_comp_2d}, while Splatter Image reconstructs the complete geometry of 3D objects with few artifacts, it struggles to generate coherent appearances with meaningful details. In contrast, our method leverages the prior knowledge embedded in a pre-trained 2D diffusion model, thereby yielding significantly improved generation quality.

% \begin{figure*}
%     \centering
%     \includegraphics[width=0.99\linewidth]{sec/images/appendix_comparisons.pdf}
%     \vspace{-0.5em}
%     \caption{\textbf{More comparisons of zero-shot text-to-3D generation.} Our generations yield better alignment with the text prompts and comparable quality to the state-of-the-art. We present the generated Gaussian atlases in the order from top left to bottom right: 3D location $\mathbf{x}$, albedo $\mathbf{c}$, color-coded opacity $\mathbf{o}$, normalized scale $\mathbf{s}$, and the last three channels of normalized quaternion $\mathbf{r}$.}
%     \vspace{-0.5em}
%     \label{fig:appendx_compare}
% \end{figure*}

\begin{figure}[t]
    \centering
    \includegraphics[width=0.99\linewidth]{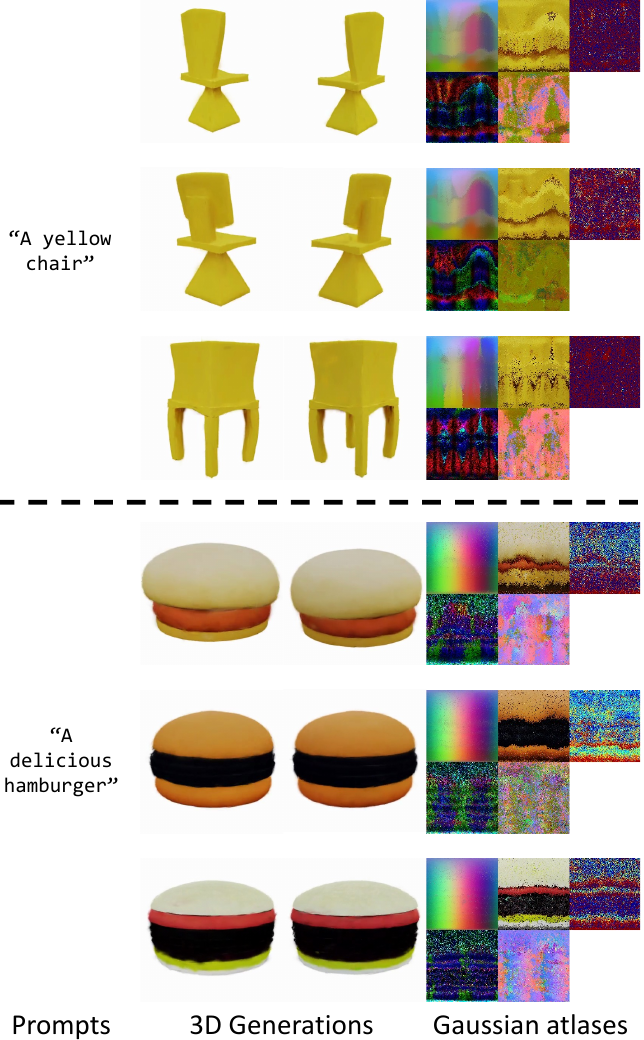}
    \vspace{-0.5em}
    \caption{\textbf{Diverse text-to-3D generation results from the same prompt.}  We present the generated Gaussian atlases in the order from top left to bottom right: 3D location $\mathbf{x}$, albedo $\mathbf{c}$, color-coded opacity $\mathbf{o}$, normalized scale $\mathbf{s}$, and the last three channels of normalized quaternion $\mathbf{r}$.}
    \vspace{-0.5em}
    \label{fig:appendx_diversity}
\end{figure}

\begin{figure}
    \centering
    \includegraphics[width=0.99\linewidth]{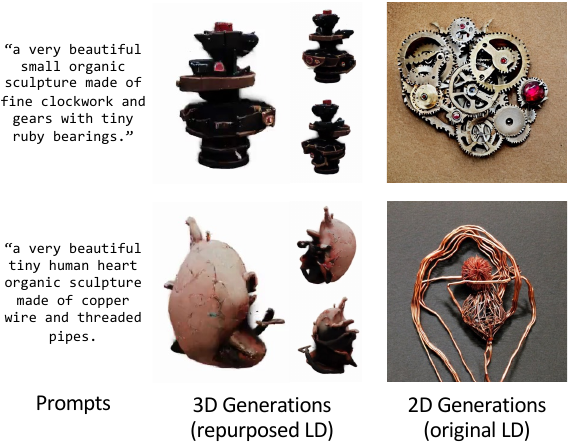}
    \caption{Similar to the original LD, our repurposed LD is not good at capturing long text prompts for 3D generations.}
    \label{fig:limitation}
\end{figure}

\begin{figure*}
    \centering
    \includegraphics[width=0.99\linewidth]{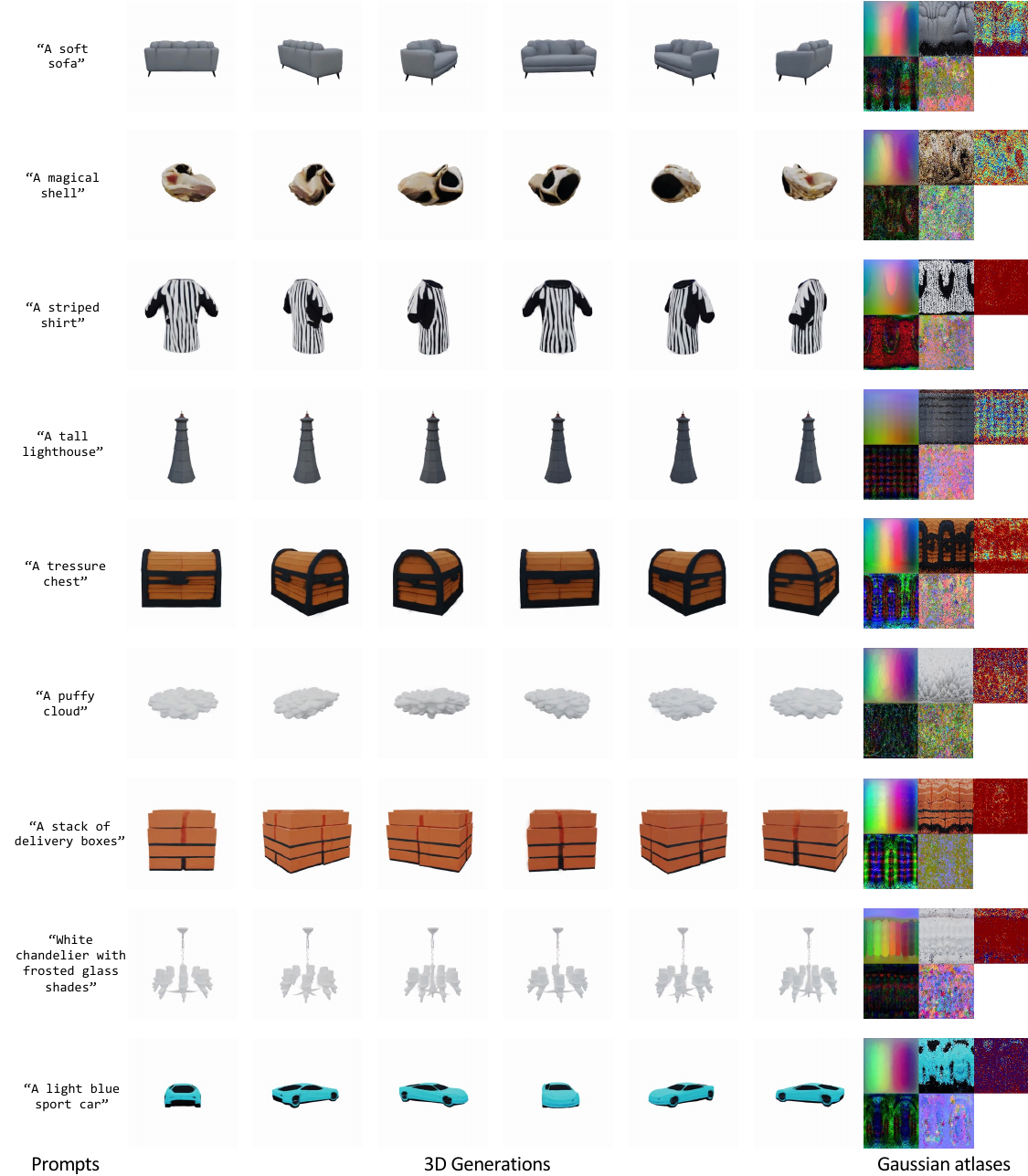}
    % \vspace{-0.5em}
    \caption{\textbf{More text-to-3D generation results.} Our model is able to generate 2D Gaussian atlases to resemble high-quality 3D Gaussians from various prompts. We present the generated Gaussian atlases in the order from top left to bottom right: 3D location $\mathbf{x}$, albedo $\mathbf{c}$, color-coded opacity $\mathbf{o}$, normalized scale $\mathbf{s}$, and the last three channels of normalized quaternion $\mathbf{r}$.}
    % \vspace{-0.5em}
    \label{fig:appendix_results}
\end{figure*}

\begin{figure*}
    \centering
    \includegraphics[width=0.99\linewidth]{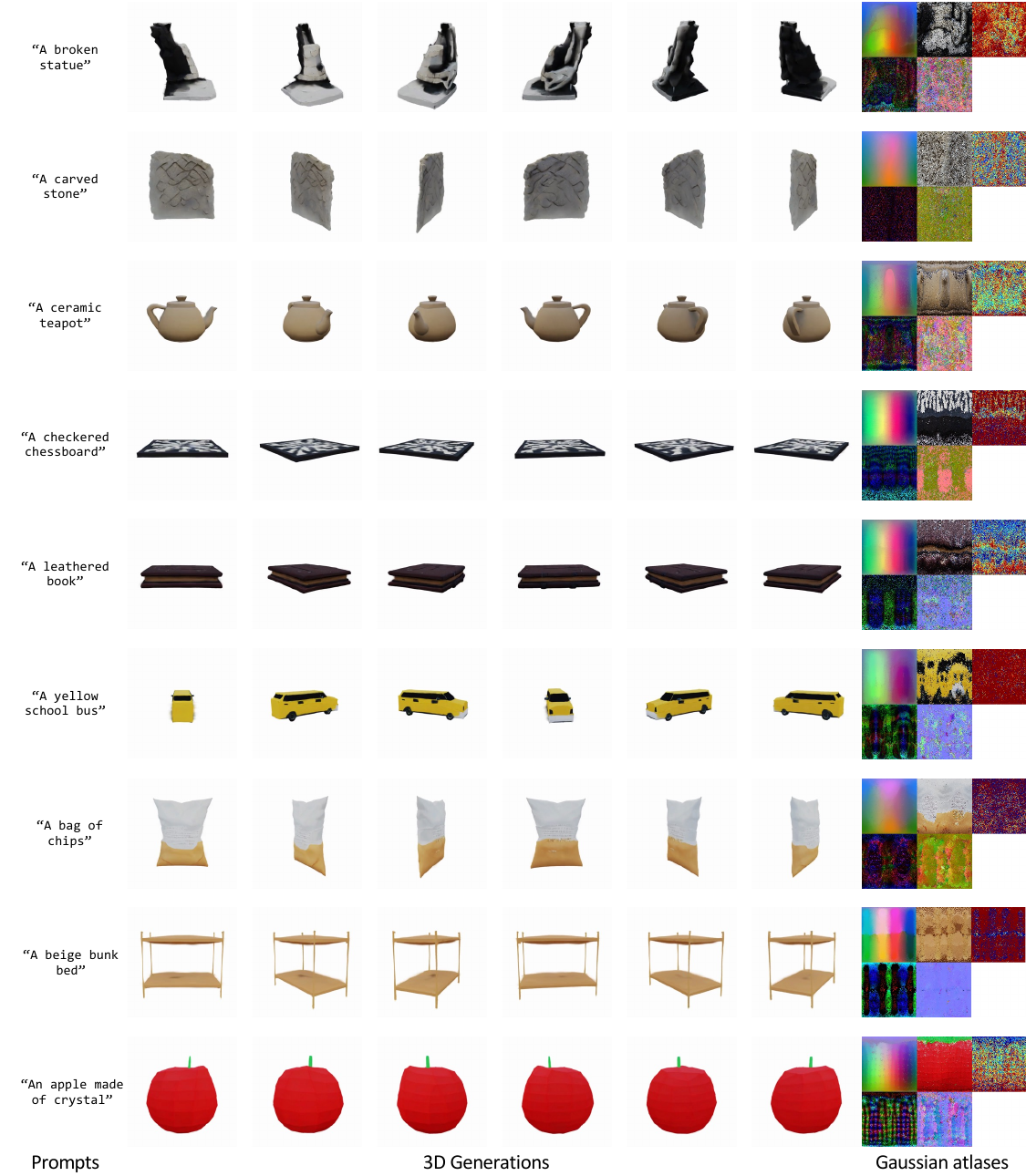}
    % \vspace{-0.5em}
    \caption{\textbf{More text-to-3D generation results (cont.).} Our model is able to generate 2D Gaussian atlases to resemble high-quality 3D Gaussians from various prompts. We present the generated Gaussian atlases in the order from top left to bottom right: 3D location $\mathbf{x}$, albedo $\mathbf{c}$, color-coded opacity $\mathbf{o}$, normalized scale $\mathbf{s}$, and the last three channels of normalized quaternion $\mathbf{r}$.}
    % \vspace{-0.5em}
    \label{fig:appendix_results2}
\end{figure*}

\subsection{More Generations}

More results generated from a diverse list of prompts are shown in \figureautorefname~\ref{fig:appendix_results} and \figureautorefname~\ref{fig:appendix_results2}. We demonstrate impressive generation quality on text prompts involving descriptions for color, shape, style, abstract semantics, and quantities for objects. Additionally, we include videos of 360-degree renderings of the generated objects along with the supplementary material.

2D flattening and surface cutting may introduce unexpected artifacts in the final 3D generations \cite{elizarov2024geometry}. However, when rendering the generated 2D atlases back as 3D Gaussians, no defects are observed at the 'seams' of disconnected 2D boundaries, and our final 3D generations are coherent and natural, similar to other direct 3D generation approaches \cite{he2024gvgen, zhang2024gaussiancube}. This proves the concept of our approach --- 3D generation achieved by a 2D diffusion model.

\subsection{Diversity of Generations}

With the same text prompt, we initiate the reverse diffusion with different random noise to demonstrate the diversity of generations in \figureautorefname~\ref{fig:appendx_diversity}. We observe a high diversity of geometries and appearances of the generated objects. This further proves that our proposed 2D representations of 3D Gaussians does not negatively alter the underlying semantics of the original 3D properties.

% More results generated from a diverse list of prompts are shown in \figureautorefname~\ref{appendix_results}. We show impressive generation quality on descriptive prompts of color, shape, style, abstract semantics, as well as quantities. We also include videos of 360 degree renderings of the generated objects along with the supplementary material.

% 2D flattening and surface cutting may raise unexpected artifacts in the final 3D generations \cite{elizarov2024geometry}. However, when rendering the generated 2D \methodname\ as 3D Gaussians, no defects are observed on the 'seams' of disconnected 2D boundaries and our final 3D generations are coherent and natural, similar to other direct 3D generation approaches \cite{he2024gvgen, zhang2024gaussiancube}. This validates the feasibility of our approach that achieves 3D generation via a 2D diffusion model.

% \subsection{Diversity of Generations}
% With the same text prompt, we start the reverse diffusion with different random noise and demonstrate the diversity of generations in \figureautorefname~\ref{fig:appendx_diversity}. We observe high diversity of object geometries and appearances. This again proves that repurposing 2D diffusion model for generating Gaussian atlases does not negatively alter the underlying semantics of 3D objects.

\section{Discussions}
\paragraph{Limitations.} Our experiments revealed a trade-off between generation quality and the parameter $N$, which denotes the number of Gaussians per Gaussian atlas. A larger $N$ allows for finer-grained details in the generated outputs, but it also increases both training and inference costs. In this paper, we set $N = 128 \times 128 = 16,384$ Gaussians as a good compromise between computational efficiency and quality. However, in our current setup — using approximately three times fewer Gaussians than in \cite{zhang2024gaussiancube} — our 3D generations do not exhibit significantly finer details compared to state-of-the-art methods that employ substantially more Gaussians.

Moreover, since this work repurposes a specific checkpoint of the LD model, no major updates have been made to the model architecture. Consequently, our model inherits the limitations of the original LD model for 3D generation, including reduced quality for long text prompts (see \figureautorefname~\ref{fig:limitation}).

%to generate highly detailed complex objects (\figureautorefname~\ref{fig:appendx_limitation}, right). Moreover, we also observed limited quality in out-of-domain generations, which may not exist in the training dataset (\figureautorefname~\ref{fig:appendx_limitation}, left).

\paragraph{Future Directions.} The purpose of this work is to introduce a novel way to represent 3D contents as 2D and make attempts to unify both 2D and 3D generation, allowing advancements in either paradigm to benefit both. There are two directions particularly worth pursuing by following the pathway presented in this work. The first is to experiment with more advanced diffusion models, such as transformer-based architectures \cite{peebles2023scalable} or latent diffusion models for Gaussians \cite{roessle2024l3dg}. The second is to integrate plug-ins that are originaly designed for 2D diffusion models, such as IP-adapter \cite{ye2023ip} and DreamBooth \cite{ruiz2023dreambooth}, into 3D generations, given the unified diffusion framework.